\documentclass[11pt,letterpaper]{article}
\usepackage[margin=1in]{geometry}
\usepackage[T1]{fontenc}
\usepackage[utf8]{inputenc}
\usepackage{newtxtext,newtxmath}
\usepackage{microtype}
\usepackage{amsmath}
\usepackage{graphicx,booktabs,tabularx,longtable,array,multirow}
\usepackage{enumitem}
\usepackage{caption}
\usepackage{float}
\usepackage{xr}
\usepackage[dvipsnames]{xcolor}
\usepackage{tikz}%
\usepackage{listings}%
\usepackage{hyperref}
\usepackage{fontawesome5}
\definecolor{releasegreen}{HTML}{1E9E5A}
\newcommand{\releaselink}[3]{\href{#1}{\textcolor{releasegreen}{#2\,#3}}}
\hypersetup{
     colorlinks = true,
     linkcolor = BrickRed,
     anchorcolor = blue,
     citecolor = Blue,
     filecolor = blue,
     urlcolor = Green
}
\usepackage[nameinlink,noabbrev]{cleveref}
\usepackage[section]{placeins}
\usepackage{fancyhdr}
\usepackage{url}
\usepackage{pdflscape}
\usepackage{needspace}
\setlist{nosep,leftmargin=*}
\renewcommand{\arraystretch}{1.12}

\newcommand{\Gemma}{Gemma-4-31B-it}
\newcommand{\Qwen}{Qwen3-32B}

\newcommand{\ahat}{\hat a}
\newcommand{\R}{\mathbb R}

\DeclareMathOperator*{\argmax}{arg\,max}
\DeclareMathOperator*{\argmin}{arg\,min}
\DeclareMathOperator{\Corr}{Corr}

\title{\vspace{-1.5em}\textbf{Jevidence: Evidence for System One Model Decisions}\vspace{-0.5em}}
\author{%
Andrew Cheng$^{1,*}$\qquad Jonathan Geuter$^{1,2,*}$\\[0.3em]
David Alvarez-Melis$^{1,2,3}$\qquad Adityanarayanan Radhakrishnan$^{4,5,6}$\\[0.65em]
{\small\begin{tabular}{@{}c@{}}
$^1$Harvard University\quad $^2$Kempner Institute\quad $^3$Microsoft Research\\[0.15em]
$^4$MIT\quad $^5$Broad Institute\quad $^6$OuterProduct, Inc.
\end{tabular}}\\[0.35em]
{\footnotesize $^*$Equal contribution}}
\date{}
\usepackage{xspace}
\newcommand{\method}{Jevidence\xspace}

\crefname{section}{Section}{Sections}
\crefname{figure}{Figure}{Figures}
\crefname{table}{Table}{Tables}
\newcommand{\appref}[1]{Appendix~\ref{#1}}

\usepackage{appendix}

\begin{document}
\maketitle\vspace{-2.2em}
\begin{abstract}
System One models (such as Jev) are emerging as an effective and efficient classifier built by fine-tuning language models.  Unlike language models, these models do not output text prior to classification, and as a result, inference is orders of magnitude faster than that of language models. Despite these advantages, System One models still lack important capabilities present in language models such as the ability to retrieve evidence (references) relevant for evaluating their output.  In this work, we introduce \method{}, a method to provide evidence for System One model outputs in a single forward pass.  Specifically, \method{} involves training and applying linear probe to predict Leave-One-Out (LOO) attribution scores from attention features for segments of input context.  As such, \method{} does not require any human-labels for producing evidence, adds little latency overhead ($2-7\%$) to System One model inference, and can be readily integrated with any open-weight System One model.  In an evaluation across System One models from the Gemma family and seven datasets, \method{} outperforms nine existing baselines in accurately recovering human-annotated evidence relevant for classification while being nearly two orders of magnitude faster than the second best method.  On the same benchmark, \method{} is also competitive with the closed-development method Locate from Perplexity while being twice as fast.  Lastly, we show that \method{} is interestingly capable of identifying evidence for correct classification even when the model produced incorrect outputs.  Together, these results show that \method{} enables fast and accurate evidence retrieval for general System One models.  
\end{abstract}
\begin{center}\small
\releaselink{https://github.com/andrewncheng/jevidence}{\faGithub}{Jevidence}\qquad
\releaselink{https://huggingface.co/datasets/andrewcheng12/jevidence-data}{\faDatabase}{jevidence-data}
\end{center}

\section{Introduction}\label{sec:intro}
System One models (e.g., Jev and its variants) are emerging as a fast and effective form of classifier built using language models~\cite{jev,opendecision2026, microsoft2026decision1, salesforce2026koa, amazon2026strandsdecider}.  Given a  question, context for answering the question, and a set of options, these models are trained to return a calibrated distribution over the options without generating any intermediate text outputs.  Unlike traditional classifiers (which typically operate on tabular data), System One models simplify predictive modeling from unstructured data such as text, with reported speedups of up to two orders of magnitude over generative LLMs on decision tasks~\cite{jev}.  

System One models are still in their nascency and do not, as of yet, contain many useful capabilities built into current language models.  One major missing capability central to advancing trust and understanding of these models is the ability to retrieve evidence (references to input context) to answer a given question.\footnote{Throughout this work, we will use the words classification, decision, and output interchangeably.} Using a language model to provide references for a System One model output is a straightforward solution, but significantly undermines the latency of System One models.  We are thus interested in a retrieval mechanism for System One models that does not sacrifice their speed and cost advantages. 

An important distinction in evidence retrieval is between identifying information that influences a model's prediction (model attribution) and information relevant to answering the question itself~\cite{worledge2023attribution}. Our interest lies in the latter. Retrieving such evidence is a well-studied task in the answer of questions and the evaluation of rationale~\cite{bohnet2022attributed,r8}. Existing approaches present a tradeoff between retrieval quality and computational efficiency. Classical attribution methods, such as Leave-One-Out (LOO)~\cite{r19}, Shapley-value estimators~\cite{r21}, and LIME~\cite{r24}, estimate input importance through repeated model evaluations. Although broadly applicable, their added latency can greatly exceed inference latency, undermining the efficiency gains of these models. Another approach is to use LLMs to produce references to supporting passages~\cite{gao2023citations}, but this approach also introduces significant latency overhead. Attention-based methods offer a more efficient alternative by leveraging features already computed during a model's forward pass~\cite{abnar2020quantifying}. Early work used attention weights to highlight important words and sentences in neural classifiers~\cite{yang2016hierarchical,mullenbach2018explainable}, while subsequent research investigated whether these weights faithfully explain model predictions~\cite{r15,serrano2019attention,r32}. Recent methods have applied this idea to model attribution~\cite{r6} and evidence retrieval~\cite{r20} in LLMs.  However, as we show, directly applying these methods to System One models yields poor evidence-retrieval performance (as compared with human annotated evidence).  

\begin{figure}[t!]
\centering
\includegraphics[width=\linewidth]{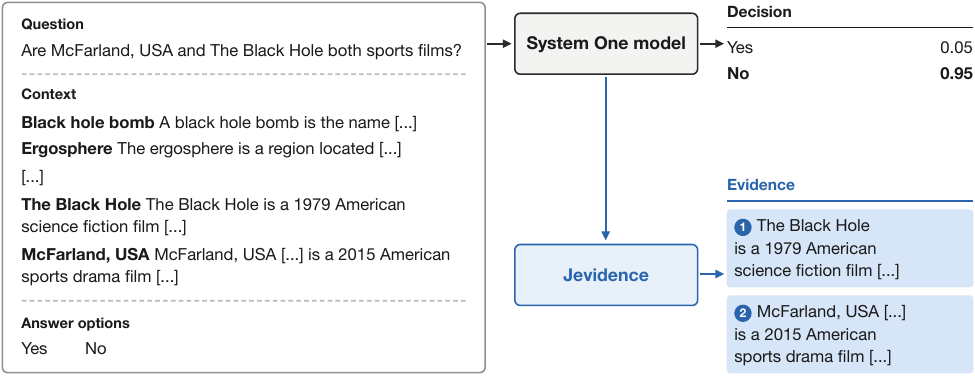}
\caption{\textbf{Jevidence enables evidence retrieval for System One models.}
A System One model receives input formatted as (question, context, possible answers) and outputs a classification from one of  the possible answers in a single forward pass.  In a parallel branch during the forward pass, Jevidence applies a pre-trained probe (on attention values) to identify context segments that provide evidence for the correct decision.}
\label{fig:format}
\end{figure}

In this work, we introduce \textit{\method{}}, a method that provides accurate evidence retrieval from System One models with nearly no additional inference cost (2-7\% overhead on inference latency).  Our approach builds on the idea of learning to extract useful information from a model's internal representations through linear probing~\cite{alain2016understanding,beaglehole2026toward} and is closely related to the attention-based attribution method of~\cite{r6}.  Specifically, we use LOO probability changes as supervision to train a linear probe on attention features.  Once trained, \method{} ranks passages in the input context using attention collected during the original model forward pass, without requiring any additional model forward passes. An overview of \method{} is presented in Figure~\ref{fig:format}.  

As our approach does not require any human annotated evidence for training, it is readily applicable to all open-weight System One models.  We apply \method{} to a broad class of 24 open weight models from under a billion parameters to up to 70 billion parameters.  Across a benchmark consisting of questions from seven different datasets, we evaluate \method{} in comparison with nine other evidence retrieval methods in its ability to accurately discover human annotated evidence relevant for decisioning.  Focusing on recently developed decision models built using Gemma models~\cite{jeff11release}, we find that \method{} outperforms all previous evidence retrieval methods in terms of retrieval accuracy (as measured by three separate metrics) while being $68-88\times$ faster than leading methods like ContextCite-128~\cite{r5} and $33-36\times$ faster than LOO.  We also find that \method{} is competitive with concurrently introduced methods.  Namely, \method{} achieves competitive evidence-retrieval performance to Perplexity's Locate function~\cite{perplexity2026PerplexityDecider} (released on Nov. 12, 2026) while being $2.2$--$2.6\times$ faster.

In addition to its effectiveness in speed and performance, our analysis of \method{} further reveals that it is robust in retrieving relevant evidence for System One model outputs unlike perturbation-based methods like LOO~\cite{r19}, ContextCite~\cite{r5}, LIME~\cite{r24}, or Kernel SHAP~\cite{r21}.
  Specifically, we show that \method{} is able to retrieve evidence for producing the right decision when (1) the decision model produces incorrect answers; (2) the input question is perturbed.  Indeed, we find that this robustness appears to be a broader property of attention-based evidence retrieval methods (e.g., AT2~\cite{r6}, SelfElicit~\cite{r20}).

Together, these results demonstrate the effectiveness and efficiency of \method{} in retrieving evidence for industry-scale System One models.

\section{Background on System One models}

System One models are classifiers built from language model backbones. Specifically, these models receive a prompt
$x=(D,q,\mathcal A)$, where $D$ denotes context data, $q$ denotes a question to be answered, and $\mathcal{A}$ denotes a finite set of possible answers.  For example, $D$ could be a passage of text, $q$ could be a multiple choice question, and $\mathcal{A}$ could denote the possible multiple choice answers. 
Given $x$, a System One model returns a probability distribution $p(a\mid x)$ where $a$ takes values in $\mathcal{A}$.  We then select an answer by selecting the value of $a$ for which $p(a \mid x)$ is largest, i.e., 
$\ahat=\argmax_{a\in\mathcal A}p(a\mid x)$.  

The idea of using LLMs as classifiers was popularized by Jev~\cite{jev} and there have since been many open-weight alternatives
~\cite{opendecision2026,liwagle2026,semif,marosi2026decider}.  In general, there are two ways to have LLMs act as classifiers: (1) by prompting vanilla LLMs with an additional  instruction to output one of the possible answers and computing $p(a \mid x)$ by normalizing the next-token logits over the permitted answers; and (2) by post-training them as classifiers directly.  We consider both types of models in this paper.  

The former only involves prompting (no post-training).  For the latter, we study models that are post-trained by the procedure introduced by~\cite{opendecision2026}, which applies LoRA with a learned decision head to open-weight LLMs.  Specifically, we consider four such models built on \Gemma{}~\cite{r10} and \Qwen{}~\cite{r23} backbones: (1) Jeff-1.1-Large (Gemma-4-31B-it), (2) Jeff-1.1-Medium (Gemma-4-12B-it), (3) Jeff-1.1-Qwen-27B (Qwen3.6-27B), and (4) Jeff-1.1-Qwen-4B (Qwen3.5-4B). We compare each post-trained model with its corresponding prompted LLM backbone, allowing us to examine how decision-specific fine-tuning (or decision tuning) affects results.  For the sake of computational efficiency, we work with models with ``thinking'' disabled.

\section{Jevidence: Evidence for System One model decisions}\label{sec:method}

We now describe \textit{Jevidence}, which is our approach for providing evidence for decisions from System One models.  In this work, we focus specifically on System One models for which the input context $D$ is text.  At a high level, given a prompt, Jevidence identifies sections of the input context $D$ that is relevant to answering the underlying question $q$. As we will observe empirically in the subsequent section, the main advantages of Jevidence are its accuracy in identifying relevant evidence and low latency overhead. 

\paragraph{Dividing context into sub-units.}  The first step in evidence retrieval is dividing up the context into sub-units that can serve as evidence.  We developed a deterministic set of segmentation rules to divide $D$ into units $u_1, \ldots, u_{m}$ where each $u_j$ can be a sentence, table row, header, etc. (Appendix~\ref{app:segmentation}).  (Note that $m$ will depend on $D$ in general.  For now, we omit this dependence to simplify the writing.) These segmentation rules were developed using open-source decision datasets and were then fixed for all datasets used in our experiments.  We note that most datasets come with their own segmentation rules, and our method can be easily adapted to such cases.  

\paragraph{Jevidence.}  To build Jevidence, we start with a given System One model $p_{\theta}$ and a dataset of $n$ input prompts, $\{x^{(i)}\}_{i=1}^n$, of the form $x^{(i)} = (D^{(i)}, q^{(i)}, \mathcal{A}^{(i)})$.  We then divide each of the datasets $D^{(i)}$ into units $(u_1^{(i)}, \ldots, u_m^{(i)})$.  The first stage to building Jevidence is to train a predictive model to identify the units in each $D^{(i)}$ that are relevant to answering $q^{(i)}$. Our supervised procedure is inspired by AT2~\cite{r6} whose goal is to identify the inputs that most influence the generative output of an LLM. In contrast, we adapt their approach to System One models and identify a different type of information, namely the evidence relevant to answering $q^{(i)}$, regardless of the model's output. This stage consists of the three steps, which we outline below.  
 
\paragraph{Step 1 (Construct training targets).}
For each input prompt $x^{(i)}$, we measure the Leave-One-Out (LOO) effect of removing the unit 
$u_j^{(i)}$ from $D^{(i)}$.  Namely, if $\hat{a} = \argmax_{a \in \mathcal{A}^{(i)}} p_{\theta}(a | x^{(i)})$, we measure:
\begin{equation}
d_j(x^{(i)})
=
p_{\theta}(\ahat\mid x^{(i)})
-
p_{\theta}(\ahat\mid x^{(i)}\setminus u_j^{(i)}).
\label{eq:loo}
\end{equation}
Using Eq.\eqref{eq:loo}, we construct the vector
$d(x^{(i)})=[d_1(x^{(i)}),\ldots,d_m(x^{(i)})]^\top\in\mathbb R^m$. Importantly, note that these targets do not require any labeled data. 

\paragraph{Step 2 (Compute attention features).}
For each selected prompt position $t$, we construct
$A_t(x)\in\mathbb R^{m\times H}$, where $H$ denotes the
number of attention heads used for attribution. Its
$(j,h)$-th entry is
\begin{equation}
A_t(x)_{j,h}
=
\sum_{s\in\operatorname{tok}(u_j)}
\alpha_h(t,s),
\label{eq:features}
\end{equation}
where $\alpha_h(t,s)$ is the attention weight from
position $t$ to token position $s$ in head $h$, and
$\operatorname{tok}(u_i)$ denotes the tokens belonging
to document unit $u_i$. In addition to this quantity, we also found that incorporating attention from
the positions immediately preceding the last prompt position $T$ was helpful for improving retrieval performance. Specifically, for
a window of $K\geq 1$ preceding positions, we define
\begin{equation}
B(x)
=
\left[
A_T(x)
\;\middle|\;
\frac{1}{K}\sum_{k=1}^{K}A_{T-k}(x)
\right]
\in\mathbb R^{m\times 2H}.
\label{eq:attention-features}
\end{equation}
Each row of $B(x)$ captures the attention that tokens pay to each document unit. Our intuition is that this quantity provides an effective initial heuristic for units that are being used as evidence for downstream evidence.  Indeed, the raw values in $B(x)$ can even be used to identify evidence directly in many cases{, as shown in the method \appref{sec:training_free} and its performance in Appendix~\ref{app:performance-results}}. Note that all features are collected during the same forward pass
that produces the decision, without additional model
evaluations (unlike many prior attribution approaches~\cite{r5,r7,r19,r24,r21}).  
$K$ is selected for each model based on a held-out validation set and criterion. We collect attention scores only from global attention layers. We provide additional details around our choices for $K$ and $B(x)$, and which attention heads to collect attention features from in \appref{app:reader}. \appref{app:at2-to-jevidence} shows the effect of these choices.

\paragraph{{Step 3 (Train the linear probe).}}
Let $b_j(x^{(i)})$ denote row $j$ of $B(x^{(i)})$.  We train a linear probe to map from $b_j(x^{(i)})$ to $d_j(x^{(i)})$.  Namely, our probe is of the form \[
s_j(x^{(i)})=\theta^{\top}b_j(x) ~;~ \theta \in \mathbb{R}^{2H}.
\]
We update $\theta$ to minimize the loss 
\[
\mathcal L(\theta)=-\frac{1}{n}\sum_{i=1}^{n}\Corr\bigl(s(x^{(i)}),d(x^{(i)})\bigr),
\qquad s(x^{(i)})=\bigl(s_1(x^{(i)}),\ldots,s_m(x^{(i)})\bigr)^{\top};
\]
where $\Corr$ denotes Pearson correlation.  (Note that $\theta$ is shared among all $j \in [m]$.)  For training, we use full-batch Adam~\cite{kingma2015adam} with a learning rate 0.03, 20 warm-up steps, and gradient clipping at 1 for at most 3,000 steps, with early stopping (patience 400 steps) on held-out data (\hyperlink{jevidence-window-selection}{Appendix~\ref*{app:reader}, ``Window selection''}). Appendix~\ref{app:reader-weights} shows the learned weights for all eight models.

\section{Empirical results}
\label{sec:evaluation}

We applied Jevidence to benchmark datasets to evaluate its effectiveness (performance and speed) in evidence retrieval for System One models.  As we will show, Jevidence is orders of magnitude faster and better aligned with human annotated evidence than existing methods.  We describe our experimental setup below.  

\subsection{Experimental setup}\label{sec:experimental_setup}
\paragraph{Models.}
We evaluate Jevidence across eight open-weight models from the
Gemma and Qwen families.  Specifically, we use four decision-tuned models
(Jeff-1.1-Large, Jeff-1.1-Medium, Jeff-1.1-Qwen-27B,
and Jeff-1.1-Qwen-4B)~\cite{jeff11release} and their
corresponding base LLMs
(Gemma-4-31B-it, Gemma-4-12B-it~\cite{gemma4report},
Qwen3.6-27B~\cite{qwen36card}, and
Qwen3.5-4B~\cite{qwen35card}). We present results for the Gemma-based models in the main text and present the Qwen model results in the appendix. To test whether \method{} generalizes, we also compare it with LOO on 16 more models (Section~\ref{sec:performance}). These include four of the strongest public System One models (StartLux~\cite{startlux2026}, basal-omni~\cite{basalomni2026}, d3~\cite{d3model2026}, and Perplexity's Decider v1.3~\cite{perplexity2026PerplexityDecider}) and twelve more base LLMs from the Gemma~4~\cite{gemma4report}, Qwen3~\cite{r23}, and Llama~3.1~\cite{llama3herd} families.

\paragraph{Baselines.}
We compare Jevidence with various attribution
methods of three types: (1) perturbation-based methods;  (2) attention-based methods; and (3) retrieval-based methods.  For perturbation-based methods, we use
leave-one-out (LOO)~\cite{r19}, ContextCite-128~\cite{r5}, LIME-32~\cite{r24} and Kernel
SHAP-32~\cite{r21}.  For attention-based methods, we use SelfElicit~\cite{r20}, AT2~\cite{r6}. Additionally, we explore simple attention pooling approaches which we call \textit{Attention (T)}, which ranks each unit by summing the attention from the last prompt token $T$ to the tokens of the unit in the spirit of classical papers \cite{r15,r32} (see the Appendix~\ref{sec:training_free} for details). For retrieval-based methods, we use BM25~\cite{r25} and random ranking.

\paragraph{Evaluation datasets.}
We considered seven 
datasets across different domains, document structures, and types of questions: MultiRC~\cite{r16,r8}, SciFact~\cite{r30}, ContractNLI~\cite{r18}, HotpotQA~\cite{r34}, FinQA~\cite{r4}, Evidence Inference~\cite{lehman2019evidence,r9}, and ConditionalQA~\cite{r28}.  Each dataset comes with a ground-truth answer and human annotations of the evidence that supports it.  To evaluate these datasets in our classification setting,
we convert each example into a standardized format
consisting of a document, a question, and
a fixed set of answer options (for an example, see Appendix~\ref{app:model-input}). 
We retain the original answer choices whenever they are provided and otherwise construct task-specific options. Across all datasets, there are a total of 3,279 questions.  We use the same context segmentation procedure across all models and attribution methods, and evaluate evidence rankings against the human annotations (Appendix~\ref{app:segmentation}).

\paragraph{Training data for Jevidence and AT2.} Among all methods, only \method{} and AT2 (Appendix~\ref{app:competing}), require a separate training dataset. We created a dataset of $44,747$ questions that were disjoint from the evaluation datasets.  Specifically 43,875 of these questions were from the public Decider released data~\cite{marosi2026decider} and the rest of the questions were on synthetic business documents collected from~\cite{opendecision2026}. (These business documents were written by Qwen3.6-27B and labeled by Qwen3.6-27B and Qwen3.5-122B-A10B.) This training data was converted to our standardized format for classification (Appendix~\ref{app:dataset_conversion}). With these inputs, we proceed with the steps in Section~\ref{sec:method} to train Jevidence (and similarly, AT2~\ref{app:competing}).  As we show below, despite being trained on short documents ($\sim$8 units per training document), Jevidence exhibits robustness to longer held-out evaluation documents ($\sim$37 units per evaluation document). (We compare \method{}'s performance across documents of various lengths in Appendix~\ref{app:length} and statistics on training and evaluation data are in Appendix~\ref{app:segmentation}) 

\paragraph{Evaluation metrics.}  We primarily evaluate methods based on whether the evidence surfaced by methods aligns with human annotations of evidence relevant for prediction.  Again, none of the methods, including Jevidence, were trained on human annotation.  This data was used solely for downstream evaluation.  Human annotations are spans of text and need not coincide with context units generated by segmentation.  We thus needed an approach for evaluating non-exact matches between context units and human annotation. 

We say that unit $u_j$ \emph{matches} an annotated span $u'$ if $|u_j \cap u'| \geq \tfrac{1}{2}\min(|u_j|, |u'|)$, where $|\cdot|$ is length in characters (details in Appendix~\ref{app:segmentation}). Consider one question whose document has units $u_{1}, \ldots, u_m$. An attribution method gives each unit a score $s_j$ and we rank the units by decreasing score. Let $E \subseteq \{1, \ldots, m\}$ be the set of indices of units that match a human annotation and let $\mathrm{top}_k$ be the indices of the $k$ highest-scoring units. We consider three metrics for evaluating methods against human annotations:
\begin{enumerate}[leftmargin=*, itemsep=3pt, topsep=4pt, parsep=0pt]
    \item \textit{Hit@$k$} $= \mathbf{1}\!\left[\mathrm{top}_k \cap E \neq \emptyset\right]$: whether at least one of the top $k$ units matches a human annotation. For $k=1$, this asks whether the highest-ranked unit does.
    \item \textit{Recall@$k$} $= |E \cap \mathrm{top}_k| \,/\, |E|$: the fraction of the matching units that appear among the top $k$.
    \item \textit{AUROC} $= \dfrac{\sum_{j \in E} \sum_{l \notin E} \mathbf{1}\!\left[s_j > s_l\right]}{|E|\,(m-|E|)}$: the probability that a random marked unit is scored higher than a random non-marked unit. Equivalently, the area under the curve plotting true positive rate against false positive rate.
\end{enumerate}
Each metric is computed per question and then averaged over the 3,279 questions. We report Hit@1, Recall@5 and AUROC in the tables, and Hit@$k$ and Recall@$k$ for $k = 1,\ldots,10$ in Figure~\ref{fig:recall-k}.

\subsection{Benchmark Results}
\label{sec:performance}

\begin{figure}[t!]
\centering
\includegraphics[width=\linewidth]{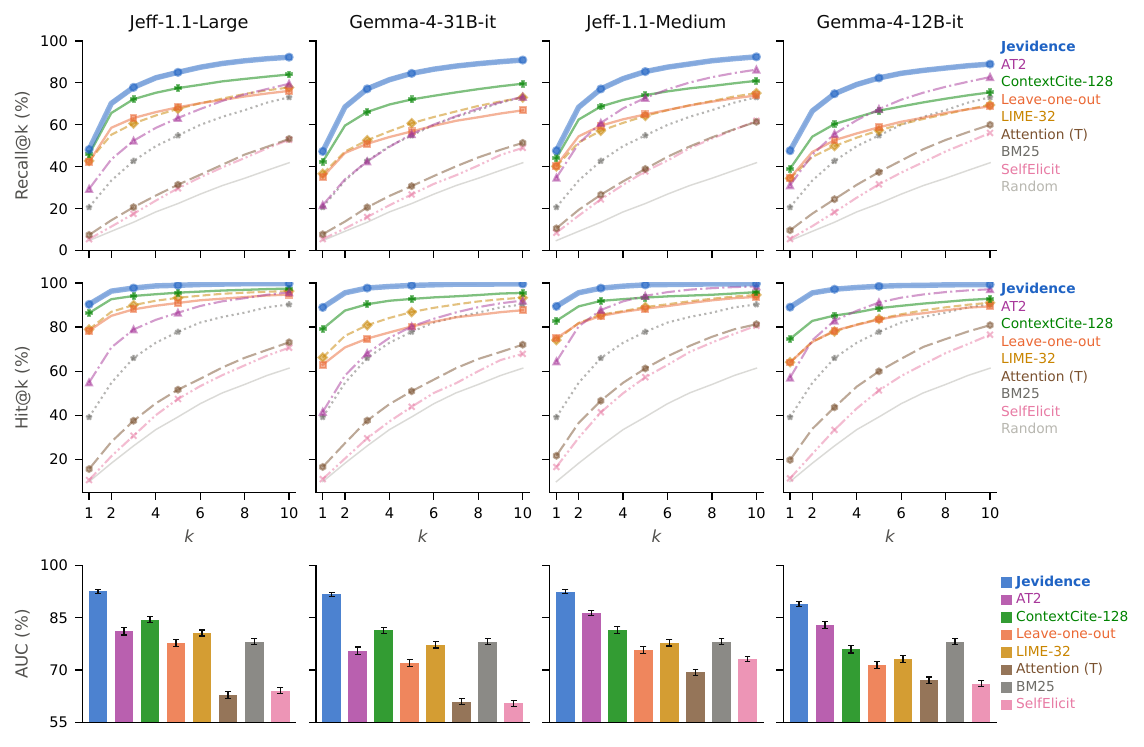}
\caption{\textbf{Evidence retrieval evaluation for System One models built using the Gemma family and their backbones.} The Random method randomly selects a subset of $k$ units. Black error bars are 95\% confidence intervals.  Appendix Figure~\ref{fig:recall-k-qwen} provides analogous results for Qwen models.}
\label{fig:recall-k}
\end{figure}

Across all datasets, System One models based on the Gemma family, and evaluation metrics, we find that \method{} is best at every $k$, the number of units retrieved (Figure~\ref{fig:recall-k}; Figure~\ref{fig:recall-k-qwen} in the appendix has the same results for Qwen models). We present a breakdown of results per dataset in Appendix~\ref{app:performance-results}. First, we observe that \method{} outperforms LOO, despite being trained exclusively on LOO targets. In \appref{app:rd-split}, we investigate this phenomenon further and find that on samples where a single unit can significantly change the model's prediction, LOO roughly ties \method{}, but on other samples (such as samples with duplicate evidence, questions where non-marked units contain evidence, or questions that the model answers without relying on the context), \method{} has improved performances.  We also find \method{} significantly outperforms AT2, even though both are attention-based methods.  We show how much each change from AT2 to \method{} benefits downstream performance in Appendix~\ref{app:at2-to-jevidence}. This ablation shows that the largest gain over AT2 comes from using attention from positions preceding $T$. 

\begin{figure}[t!]
\centering
\includegraphics[width=\linewidth]{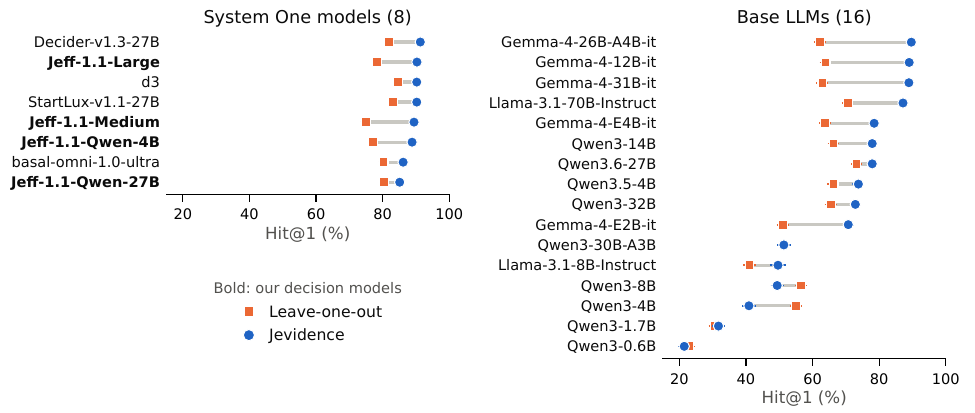}
\caption{\textbf{\method{} vs. LOO on 24 LLM-based classifiers.}  Hit@1 over the 3,279 marked questions for LOO (orange squares) and \method{} (blue circles), with 95\% confidence intervals. The left panel shows System One models, the right panel shows base LLMs prompted for classification. Rows are sorted by \method{}'s Hit@1. Each model's \method{} is trained on that model's own LOO scores, with $K$ selected on the development set.}
\label{fig:generalization}
\end{figure}

Given that \method{} does not require any human annotated labels, it can be readily applied to any open weight System One model. Figure~\ref{fig:generalization} compares \method{} with LOO on all 24 models of Section~\ref{sec:experimental_setup}.  \method{} has a higher Hit@1 than LOO on 20 of the 24 models, including all four outside decision models. The four exceptions are Qwen3 base LLMs (0.6B, 4B, 8B, and 30B-A3B).

\begin{figure}[t!]
\centering
\includegraphics[width=\linewidth]{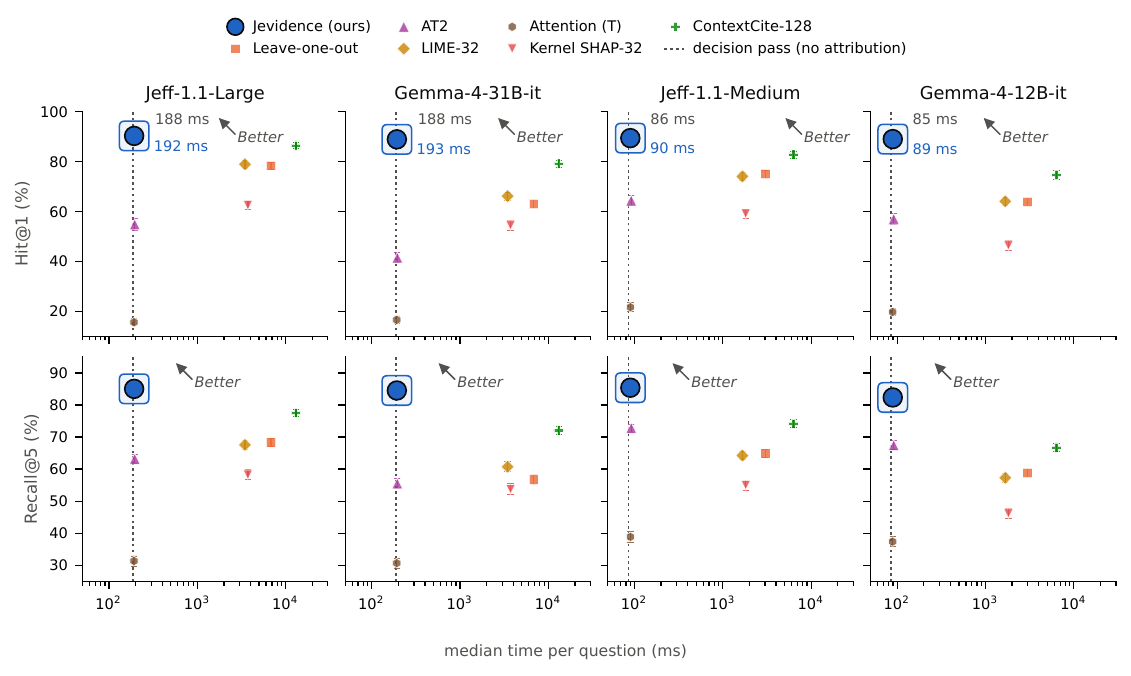}
\caption{\textbf{Performance vs. median inference latency for classifiers based on Gemma.} Hit@1 (top) and Recall@5 (bottom) over all 3,279 marked questions against the median time per question over the same questions at batch size one. The dotted line indicates inference latency without evidence retrieval. The blue circle indicates the same pass with \method{} applied. {Numbers in the top row are the median inference times (grey) and median inference time with evidence from \method (blue).} The overhead comes from mapping the document's units to tokens ($\sim$3.5~ms) and extracting the attention ($\sim$2~ms) which contributes to the {2--7}\% above the vanilla decision pass. Results for classifiers built on Qwen models are shown in Appendix Figure~\ref{fig:latency-qwen}.}
\label{fig:latency}
\end{figure}

Beyond performance, a major advantage of \method{} over other methods is that its evidence retrieval requires no additional System One model forward passes.  In contrast, methods like ContextCite-128 require up to $128$ additional forward passes. We showcase this difference in Figure~\ref{fig:latency}, where we compare evidence retrieval quality against inference latency across the 3,279 questions, measured on a single H200 GPU.    \method{} far outperforms other methods on both retrieval and inference and adds only 2-7\% to the median inference latency of the base decision model.  (Appendix~\ref{app:latency-mean} reports the corresponding
comparison using \emph{mean} inference latency, which also shows performance and speed gains of \method{} over other methods.) LOO is 33 to 36 times slower than \method{}, and ContextCite-128 is 68 to 88 times slower. We show Jevidence on five questions from each of the evaluation datasets in Appendix~\ref{app:example-outputs}.

\subsection{Comparison between Jevidence and Perplexity's Locate}\label{sec:locate}
Perplexity's Decider v1.3 extended their System One models with a new function \emph{Locate}, which identifies regions of text relevant for answering a question~\cite{perplexity2026PerplexityDecider}. Locate introduces a separate LoRA adapter on top of a fine-tune of Qwen3.8-27B, as well as a \textit{span head}.
It also introduces a special \texttt{<|locate|>} token with a learned embedding. The span head combines the \texttt{<|locate|>} representation with input-token representations to score span starts, ends, and their pairings, and predicts field-specific thresholds for selecting spans. Training uses ground-truth spans from named entity recognition, personal information detection, and extractive question answering, with an adapted InfoNCE objective that pushes correct span scores above the thresholds and incorrect span scores below them. The LoRA adapter and span head are trained while keeping the backbone frozen.
Below, we first review key differences between \method{} and Locate.  Then, we compare the two on our benchmark.

\paragraph{Differences between Jevidence and Locate.}
Jevidence and Locate operate at different levels of
granularity.  Jevidence attributes predefined context
units (e.g., sentences, table rows, and headers), whereas
Locate extracts variable-length text spans, potentially
as short as a single word or number. To enable a common
comparison on our evaluation datasets, we map each Locate span to the unit with which it has the greatest character overlap and evaluate the resulting unit rankings against human-annotated evidence. This conversion can credit Locate with an entire unit even when it extracts only a small portion of that unit. 

The methods also differ in their training supervision.
\method{} does not require any human-annotated data for training and instead trains on LOO targets, whereas Locate requires ground-truth spans for training. We also note that \method{} has been trained on multiple-choice questions, whereas Locate appears to only have been trained on question-answering; our experiments show that it also generalizes quite well to multiple-choice style extraction.

\begin{figure}[t!]
\centering
\includegraphics[width=\linewidth]{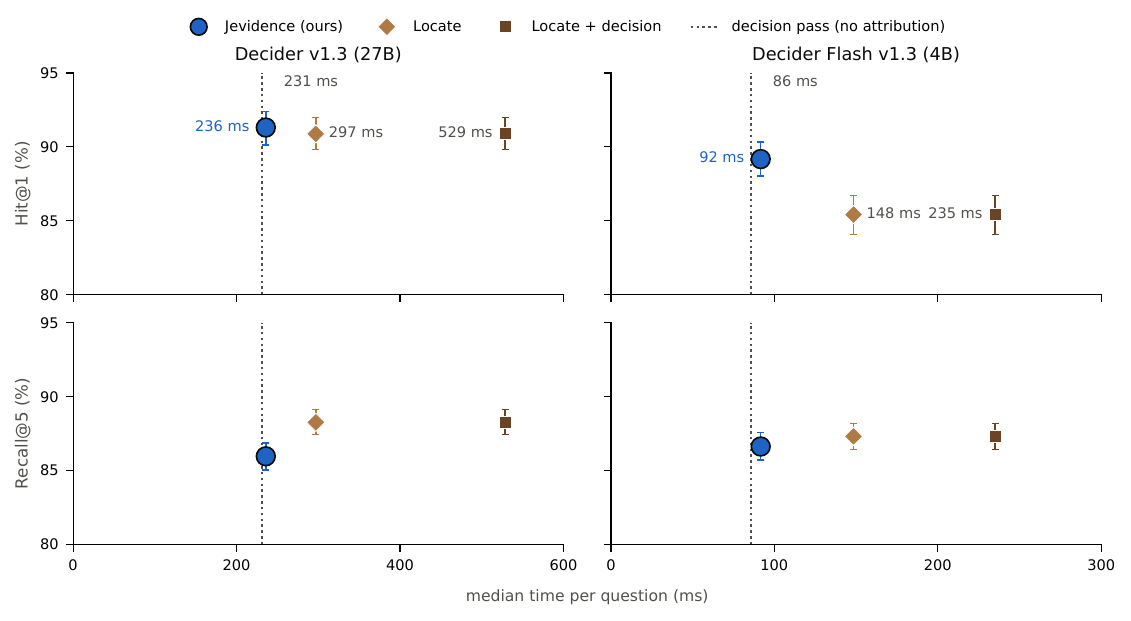}
\caption{\textbf{\method{} vs Perplexity's Locate.} Hit@1 (top) and Recall@5 (bottom) over the 3,279 marked questions against the median time per question. \emph{Locate} measures the latency of evidence extraction alone, whereas \emph{Locate + Decision} includes both prediction and evidence extraction, which we run sequentially. For each model, we report Locate's best performance on the evaluation sets across seven prompt variants~(Appendix~\ref{app:locate}). The dotted vertical line indicates inference latency of the decision pass alone, and vertical error bars denote 95\% confidence intervals.}
\label{fig:locate-latency}
\end{figure}

Lastly, \method{} returns the decision and evidence attribution in a single forward pass, while Locate instead separates them into two independent forward passes (see Figure~\ref{fig:locate-latency}).

\paragraph{Comparison between \method{} and Locate on our benchmark.}  We compared Jevidence to Locate on Decider models (Decider-v1.3-27B and Decider-Flash-v1.3-4B), where trained Locate weights exist. To adapt Locate to our multi-class classification setting, we evaluated seven prompt variants that differed in instruction wording (including instructions suggested by the Locate authors).  For each model, we select the prompt variant resulting in Locate's best performance on the evaluation data (details in Appendix~\ref{app:locate}). Locate is trained to output a score for each candidate span of tokens and returns spans with scores above a fixed threshold. In our experiments, we set this threshold to zero and return Locate's top $k$ spans. Indeed, the default threshold of 0.5 did not return any spans for most questions (76.2\% of questions on Decider, and {71.6}\% on Decider Flash).

Figure~\ref{fig:locate-latency} and Table~\ref{tab:locate} compare Jevidence and Locate in terms of latency and retrieval performance across the 3,279 questions in our benchmark.  A breakdown of results per dataset in our benchmark is available in Table~\ref{tab:locate-sets}.  In all cases, we found that \method{} (with both evidence and classification) has lower latency than Locate alone. An additional decision pass on Decider adds to the latency (where we report latency from running both forward passes sequentially, but we note that they are independent and could be run in parallel). In all cases, we also found that \method{} and Locate have comparable retrieval accuracy (within a few percentage points across all metrics in all cases except for Hit@1 for Decider Flash, where \method{} has roughly a {$3.8$-point} lead). 

\subsection{{Robustness of \method{}}}\label{sec:finqa-cf}

Beyond evaluating performance of \method{}, we sought to evaluate the robustness of \method{} in retrieving relevant evidence.  We aimed to evaluate \method{}'s robustness with respect to two settings: (1) when the base System One model is inaccurate in classification; and (2) when the input prompt formatting changes.  

\begin{figure}[t!]
\centering
\includegraphics[width=\linewidth]{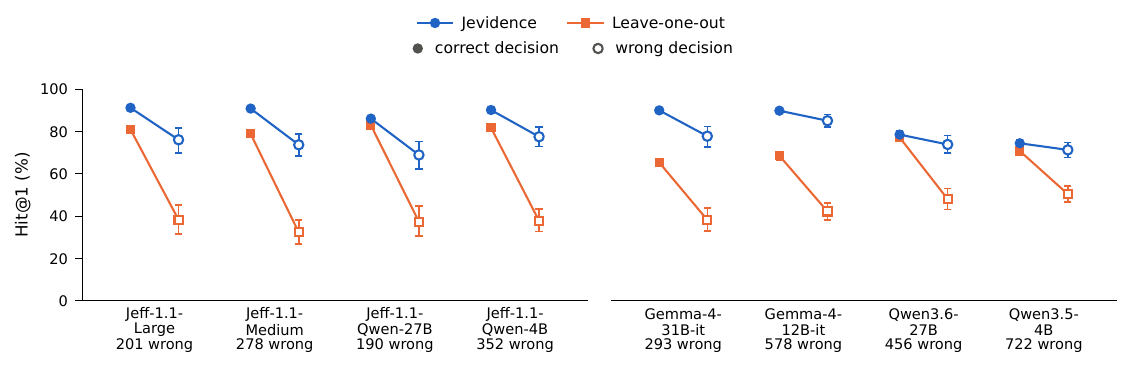}
\caption{{\textbf{Evidence retrieval stratified by model accuracy.} Hit@1 of \method{} (blue) and LOO (orange) on the 3,279 marked questions, split by whether the model's decision matches the dataset's answer (correct (filled marker) or wrong (open marker)). The number under each model indicates the number of the questions it answers incorrectly.}}
\label{fig:wrong-decisions}
\end{figure}

Through our experiments, we first found that \method could retrieve relevant evidence for correctly answering a question even when a proposed answer is incorrect.  Moreover, this finding did not appear to hold for LOO.  To illustrate this phenomenon, we stratified the 3,279 evaluation questions by whether the model was correct or not and then stratified by the Hit@1 performance for \method{} and LOO (Figure~\ref{fig:wrong-decisions}).  Overall, we find that \method{} is more accurate than LOO in retrieving accurate evidence for answering the question than LOO, regardless of whether the answer produced by the decision model was correct or incorrect.  Appendix~\ref{app:wrong-decisions} repeats the same experiment but for Recall@5 and AUROC and Table~\ref{tab:wrong-decisions-per-dataset} shows Jevidence's performance on correct/incorrect questions across datasets and models.

\begin{figure}[t!]
\centering
\includegraphics[width=\linewidth]{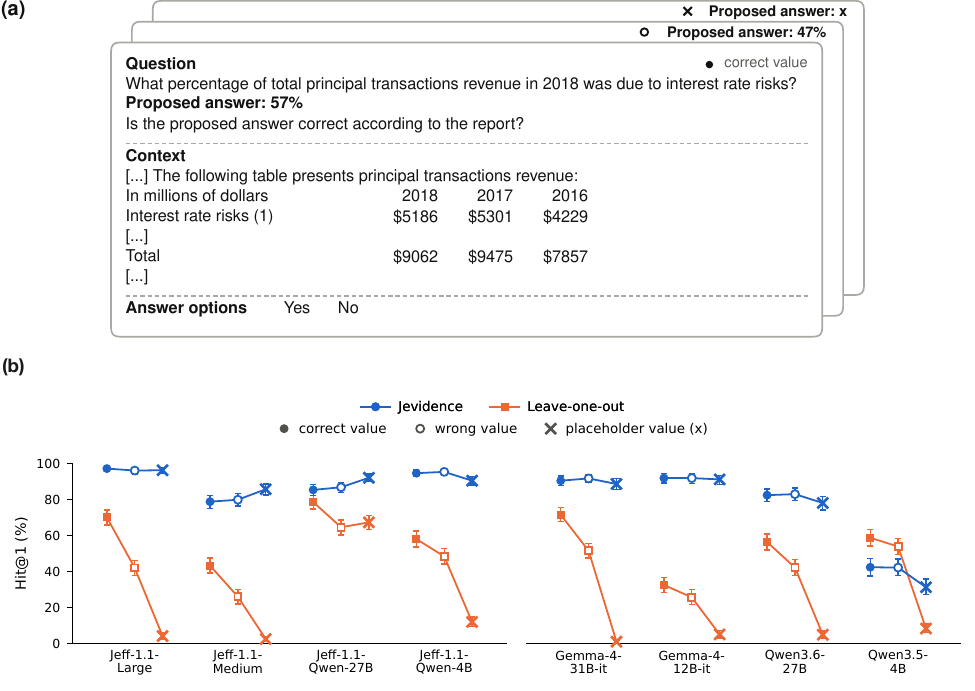}
\caption{\textbf{Evidence retrieval robustness to prompt changes.} (a) Fora given context, and answer options, FinQA-CF provides three variations changing the question: (i) correct answer (57\%); (ii) wrong answer (47\%); (iii) the placeholder (the letter \texttt{x}). (b) Hit@1 on FinQA-CF for \method{} (blue) and LOO (orange) stratified by prompt in (a).  Error bars show 95\% confidence
intervals. 
}
\label{fig:finqa-cf}
\end{figure}

We then evaluated the robustness of \method{} to prompt formatting.  To this end, we considered a task where a System One model receives a proposed answer as part of the question and is asked to decide whether the proposed answer is correct or not.  The goal was to determine whether \method{} retrieves the necessary evidence for answering the question correctly, regardless of the value of the proposed answer (i.e., a systematic change in the prompt).  To this end, we built FinQA-CF, a variant of FinQA consisting of 550 base inputs from FinQA.  We modify each input to produce three new  model inputs where we introduce a proposed answer that takes on one of the following values: 
\begin{enumerate}[leftmargin=*, itemsep=1pt, topsep=3pt]
    \item \emph{Correct value:} The true answer.
    \item \emph{Wrong value:} A wrong answer generated by redoing the calculation required for the true answer but replacing one of the steps by a neighboring number from the report (such as the same line item for a different year).
    \item \emph{Placeholder:} Inserting the letter \texttt{x}.
\end{enumerate}
Figure~\ref{fig:finqa-cf}a presents an example of these three inputs from FinQA-CF.  In \Cref{fig:finqa-cf}b, we evaluate how \method{} and LOO respond when the proposed value changes.  Overall, we observe that \method{} is barely affected by changing the proposed value.  On the other hand, LOO performs noticeably worse (up to a $60\%$ or larger drop in cases) when the placeholder or wrong value is introduced. 
Appendix~\ref{app:finqa-cf-methods} includes results for other attribution methods in this setting.  We generally find that attention-based methods are less sensitive than perturbation-based methods to changes in the proposed answer, suggesting why \method{} (a model trained on attention) may be less sensitive than LOO.  

\section{Conclusion}
By fine-tuning language models for classification, System One models (popularized by Jev) enable fast and accurate predictive modeling at scale. Yet, these models are still early on in development and lack many capabilities of current language models.  In this work, we developed \method{}, a method to enable accurate evidence retrieval (references) from open-weight System One models with nearly no additional latency overhead.  Specifically, \method{} involved training a linear probe to predict LOO probability changes from attention features.  As it does not require any human-annotated data for training, \method{} can be readily applied to any open-weight System One model. We demonstrated its effectiveness and speed relative to existing evidence retrieval methods and attribution methods across models ranging from less than one billion parameters to up to 70 billion parameters.  

Beyond benchmark evaluations, we found that \method{} could interestingly identify relevant evidence for classification even when a System One model's output was incorrect. This phenomenon suggests a promising downstream application of developing fast feedback loops for System One models in which retrieved evidence is used to assess model decisions, and when necessary, help correct them. Our work also opens several directions for future research. For example, although \method{} uses a simple linear probe on attention features, more expressive probes may further improve evidence retrieval without sacrificing efficiency (in line with existing probing findings from~\cite{beaglehole2026toward}).  

More broadly, our results suggest System One models efficiently and robustly support evidence retrieval.  By providing evidence alongside predictions, \method{} expands the range of applications for System One models, particularly in regulated industries where evidence retrieval is important for model governance.  We envision this capability is key to enabling scalable, transparent predictive AI.

\subsection*{Acknowledgments}
We would like to thank Bobak Kiani for helpful discussions, as well as Jonas Golde and Bo Wang from Perplexity for their insights on Locate.  AC is supported by the NSERC postgraduate scholarship.
JG is
supported by a Kempner Graduate Fellowship. DAM acknowledges support from the Kempner Institute and NSF award no.~2229881.  

\section*{AI Statement}
The authors used AI agents to help with writing code, running experiments, and writing. AI-generated content was thoroughly verified, and the authors take responsibility for the final contents of this work.

\clearpage
\appendix
\makeatletter
\let\apx@addcontentsline\addcontentsline
\def\apx@toc{toc}
\renewcommand{\addcontentsline}[3]{\def\apx@file{#1}%
  \ifx\apx@file\apx@toc\apx@addcontentsline{apx}{#2}{#3}%
  \else\apx@addcontentsline{#1}{#2}{#3}\fi}
\section*{Appendix contents}
{\hypersetup{linkcolor=black}\@starttoc{apx}}
\makeatother
\clearpage
\section{Methods and data}\label{app:methods-data}

This section describes \method{}, and a training-free evidence-retrieval approach that we designed and achieves competitive performance on Gemma~4 backbones (Appendix~\ref{sec:training_free}). We also provide details for competing attribution methods.

\subsection{\method{}}\label{app:reader}

\paragraph{Inputs and outputs.}
Consider an input $x=(D,q,\mathcal A)$, where the
document $D$ is partitioned into $m$ units
$u_1,\ldots,u_m$, and $\mathcal A$ is the set of
available answer options. Let $T$ be the number of
tokens in the full prompt and $d$ the hidden
dimension of the decision model. The input token embeddings form a matrix
$X^{(0)}\in\R^{T\times d}$. The frozen transformer,
consisting of $L$ layers, produces final hidden
representations $X^{(L)}\in\R^{T\times d}$.
The final position $h_T=X_T^{(L)}\in\R^d$
determines the decision distribution
$p(\cdot\mid x)\in\R^{|\mathcal A|}$. \method{} uses attention weights collected during
this same forward pass to assign an attribution
score to each document unit. Its input is a matrix
of unit-level attention features, and its output
is a vector $s(x)\in\R^m$ containing one score
per document unit.

\paragraph{Features and score.}
Let $H$ denote the total number of selected
attention heads across the transformer's
global-attention layers. Unit $u_i$ is represented by row $i$ of $B(x)$ (\cref{eq:attention-features}): the attention that each of the $H$ heads of the global-attention layers gives to the tokens of $u_i$ from the finaly prompt position $T$, and its mean over the $K$ positions before $T$,
\[
b_i(x)=\Bigl(\underbrace{A_T(x)_{i,1},\ldots,A_T(x)_{i,H}}_{\text{readout position }T},\ \underbrace{\bar A_K(x)_{i,1},\ldots,\bar A_K(x)_{i,H}}_{\text{mean over }T-1,\ldots,T-K}\Bigr)^{\!\top}\in\R^{2H},
\qquad
\bar A_K(x)=\frac{1}{K}\sum_{k=1}^{K}A_{T-k}(x).
\]
The probe gives each feature one weight and ranks the units by decreasing score
\[
s_i(x)=\theta^{\top}b_i(x)=\sum_{h=1}^{H}\theta_h\,A_T(x)_{i,h}+\sum_{h=1}^{H}\theta_{H+h}\,\bar A_K(x)_{i,h},
\qquad\theta\in\R^{2H}.
\]
It has no bias term, because the loss below compares scores only within a question, where a common shift changes neither the loss nor the ranking; it thus has $2H$ parameters, (see~\cref{tab:reader-spec} for parameter counts across models).

\paragraph{Target and fit.}
The targets are the leave-one-out drops $d(x)$ of \cref{eq:loo}, and $\theta$ minimizes
\[
\mathcal L(\theta)=-\frac{1}{|\mathcal D|}\sum_{x\in\mathcal D}\Corr\bigl(s(x),d(x)\bigr),
\qquad s(x)=\bigl(s_1(x),\ldots,s_m(x)\bigr)^{\top},
\]
over the 44,747 training questions $\mathcal D$, with $\Corr$ the Pearson correlation across a question's units (questions with constant drops are left out). The features are divided by their root mean square over the 379,405 training units, and $\theta$ starts from equal weights on the raw features (their plain sum) and is fitted by full-batch Adam (learning rate 0.03, 20 warm-up steps, gradient clipping at 1) for at most 3,000 steps, with early stopping (patience 400 steps) on the development criterion below. The fit is deterministic, so the three training seeds of \cref{sec:experimental_setup} give the same probe.

\paragraph{Selecting attention heads with signal.} 
For all attention-based methods (\method{}, AT2, training-free methods~(Appendix~\ref{sec:training_free})) we collect attention features only from \emph{global-attention layers}: layers in which a token can attend to every earlier token. For example, for Gemma~4 models these are the \texttt{full\_attention} layers. We ignore Gemma's \texttt{sliding\_attention} layers, which are the sliding-window attention layers where a query can attend to only the preceeding $w=1024$ tokens. For Qwen models, we focus only on \texttt{full\_attention} layers and ignore the \texttt{linear\_attention} layers. Our experiments found that attention-based methods perform better if we collect attention scores only from the global attention layers. Hence, all our attention-based experiments collect attention scores from all global attention layers, and ignores any other layers. For reference, Gemma-4-31B-it has 10 global layers and $H=320$ heads and Qwen3.6-27B has 16 global attention layers and $H=384$ heads. We leave a systematic study of how one should select a subset of attention layers for future work.

We created a dataset of $44,747$ questions that were disjoint from the evaluation datasets.  Specifically 43,875 of these questions were from the public Decider released data~\cite{marosi2026decider} and the rest of the questions were on synthetic business documents collected from~\cite{opendecision2026}. (These business documents were written by Qwen3.6-27B and labelled by Qwen3.6-27B and Qwen3.5-122B-A10B.)

\paragraph[Window selection.]{\hypertarget{jevidence-window-selection}{Window selection.}}
The window length $K$ is the number of prompt positions before $T$ whose attention features are used as input in to \method{} (\cref{eq:attention-features}). We select $K$ separately for each model, on held-out validation questions, as the value that \emph{maximizes} the criterion $Q(K)$ defined below. 

We measure this on 2,320 validation questions, held out from the same two sources as the training questions: 2,215 from the public Decider tasks~\cite{marosi2026decider}, and 105 from the smaller source that includes the synthetic business documents of~\cite{opendecision2026}. The validation dataset is disjoint from the training and evaluation datasets. For a validation question $x$, let $d_i(x)$ be the leave-one-out drop of unit $i$  (Equation~\eqref{eq:loo}) and let $\hat\imath=\argmax_i s_i(x)$ be the unit that \method{} ranks first. For a question $x$,
\[
q(x)=\frac{\max\bigl(0,d_{\hat\imath}(x)\bigr)}{\max_i d_i(x)}\in[0,1]
\]
is the share of the largest leave-one-out drop that this unit recovers: $q(x)=1$ when \method{}'s first pick is
the unit whose removal lowers the answer probability the most, and $q(x)=0$ when removing its pick does not
lower that probability at all. Higher is better. We use only questions whose largest drop is at least $0.05$,
that is, questions where removing some unit visibly changes the answer. $Q(K)$ is the mean of $q$ over the
questions of each group, averaged over the two groups, so that the smaller group counts as much as the
larger one. For each model we fit one probe for every $K\in\{1,3,5,6,7,8,9,11,13,15\}$ and take
$K^*=\argmax_K Q(K)$; \cref{tab:reader-spec} lists the selected values. 

\begin{table}[!htbp]\centering\small\setlength{\tabcolsep}{6pt}
{
\begin{tabular}{lrrr}\toprule
Model & Heads read, $H$ & Window, $K^*$ & Parameters, $2H$\\\midrule
\multicolumn{4}{l}{\emph{Post-trained models}}\\
Jeff-1.1-Large & 320 ($10\times32$) & 11 & 640\\
Jeff-1.1-Medium & 128 ($8\times16$) & 13 & 256\\
Jeff-1.1-Qwen-27B & 384 ($16\times24$) & 8 & 768\\
Jeff-1.1-Qwen-4B & 128 ($8\times16$) & 8 & 256\\
\addlinespace[3pt]
\multicolumn{4}{l}{\emph{Base LLMs}}\\
Gemma-4-31B-it & 320 ($10\times32$) & 3 & 640\\
Gemma-4-12B-it & 128 ($8\times16$) & 15 & 256\\
Qwen3.6-27B & 384 ($16\times24$) & 9 & 768\\
Qwen3.5-4B & 128 ($8\times16$) & 11 & 256\\
\bottomrule\end{tabular}
}
\caption{{\textbf{\method{}'s specification across models.} $H$ counts the heads of the global-attention layers (layers $\times$ heads per layer; \cref{sec:method}), and $K^*$ is the window selected on the development questions. The probe has one weight per feature and no bias, so $2H$ parameters.}}
\label{tab:reader-spec}
\end{table}

\subsection{Training-free attribution}\label{sec:training_free}

We consider two simple attribution methods that require
no training. Both rank document units by aggregating
attention scores collected during the System One model's
forward pass.

\paragraph{Final prompt position.}
Our first method sums the attention that the final
prompt position $T$ assigns to each document unit
across all global-attention heads:
\begin{equation}
s_i^{\mathrm{free}\text{-}T}(x)
=
\sum_{h=1}^{H} [A_T(x)]_{i,h}.
\end{equation}
Here, $H$ is the number of global-attention heads,
and $[A_T(x)]_{i,h}$ is the attention that head $h$
assigns to document unit $u_i$ from position $T$.
We rank units by decreasing score. This is the \textit{Attention (T)} baseline from Section~\ref{sec:experimental_setup}, and is the closest baseline to the setting considered in the classical papers of \cite{r15,r32}.

An alternative formulation that we introduce is incorporating the $K$ positions immediately preceding the final input position $T$ {as well}:
\[
s_i^{\mathrm{free}}(x)
= {\frac{1}{K+1}}\sum_{j=0}^{K}\sum_{h=1}^{H}[A_{T-j}(x)]_{i,h}.
\]
We rank units by decreasing score. In the reported experiments, $K=3$, so the score uses attention from {$T$, $T-1$, $T-2$, and $T-3$}. The method uses the same global attention heads as \method{} and requires only the forward pass that produces the decision. It requires neither training nor LOO targets.

Results for both scores are reported in Appendix~\ref{app:performance-results}.

\subsection{Competing methods}\label{app:competing}
In this section we provide details to LOO, ContextCite-128, LIME-32, Kernel SHAP-32, SelfElicit, BM25, Random{, AT2 and Attention (T)}, i.e. our competing baselines.

\paragraph{Leave-one-out (LOO).}
Leave-one-out measures the influence of each
document unit on the model's prediction by
removing that unit and recomputing the decision
probabilities. Let $p(a\mid x)$ denote the frozen decision
model's probability assigned to option $a$, and
let $\hat a=\argmax_{a\in\mathcal A}p(a\mid x)$
be its selected answer on the full input. For each unit $u_i$, we construct a perturbed
input $x\setminus u_i$ by deleting $u_i$ from
the document while keeping the question and
answer options unchanged. The attribution
score is
\begin{equation}
d_i(x)
=
p(\hat a\mid x)
-
p(\hat a\mid x\setminus u_i).
\end{equation}

The original prediction $\hat a$ is held fixed
across all deletions. The method outputs
$d(x)=(d_1(x),\ldots,d_m(x))^\top\in\R^m$,
and ranks document units by decreasing score.
A positive score indicates that removing the
unit reduces the probability assigned to the
original prediction. LOO requires $m+1$ forward passes per question:
one on the original input and one for each of
the $m$ document-unit deletions.

\paragraph{AT2.}
Attribution with Attention (AT2)~\cite{r6} scores each input token by a
weighted combination of attention heads, with the weights trained to predict
how the model's output changes when random parts of the input are removed.
Once trained, it needs only the attention of a single forward pass.

AT2 was designed for generated responses and aggregates
attention from the response positions. Our decision models output option
probabilities without generating a response, so we read attention from the
final prompt position $T$. For an input $x=(D,q,\mathcal A)$ with
document units $u_1,\ldots,u_m$, let $A_T(x)\in\R^{m\times H}$ denote the
unit-level attention features at position $T$, where $H$ is the number of
attention heads. AT2 learns a weight vector $\theta\in\R^H$ and assigns each
unit the score
\begin{equation}
s_i^{\mathrm{AT2}}(x)
=
\sum_{h=1}^{H}
\theta_h A_T(x)_{i,h}.
\end{equation}

We train AT2 on the same 44,747 questions as \method{},
following the recipe of the authors' released code. For each question we draw
32 random masks $z$, each keeping every document token independently with
probability $\tfrac12$, and evaluate the frozen decision model on each masked
input $x_z$. The training target is the log-odds of the original predicted
answer $\hat a$,
\begin{equation}
y(z)
=
\log\frac{p(\hat a\mid x_z)}
{1-p(\hat a\mid x_z)},
\end{equation}
and the head weights are fitted to maximize, across the masks of a question,
the Pearson correlation between $y(z)$ and a linear surrogate that sums the
attention-based scores of the retained tokens. Optimization uses Adam with
learning rate $10^{-3}$ and a cosine schedule for 1,000 steps at batch size
512, with gradient clipping at 1 and no early stopping or selection on
development data.

\emph{Differences from the original.} We make two changes from the original paper~\cite{r6} to improve their performance on our tasks and report the best result for AT2 that we can find. First, masked
tokens are removed from the input rather than hidden from attention. Hiding tokens from tokens from attention works by blocking attention weights, which the linear-attention layers of Qwen models do not have, so on those models the hidden token would still influence the output. Removing these tokens hides them from every layer

Second, the original AT2 weights every head in every layer, whereas we fit it on the global-attention layers. Fitted on all heads as published in \cite{r6}, AT2 performs poorly on the four
Gemma-architecture models: its Hit@1 is 38.2 on Jeff-1.1-Large, 43.3 on
Jeff-1.1-Medium, 27.7 on Gemma-4-31B-it and 34.8 on Gemma-4-12B-it.
Restricting it to the global heads raises these to 54.9, 64.4, 41.5 and 57.1,
a gain of 13.9 to 22.2 points. We report the strongest performance for AT2.
Apart from these changes, our implementation is a port of the authors' code and reproduces it:
on Jeff-1.1-Qwen-4B, where no head-set choice is involved, the authors' own
training loop yields the same head weights and the same top-ranked unit on
every evaluation question.

At inference, the learned weights are applied to the attention recorded
during the original decision pass, so AT2 costs no additional model
evaluations. Unlike \method{}, which combines attention from $T$ and the
preceding prompt positions and is trained on leave-one-out targets, AT2 reads
only position $T$ and learns from random masking.

\paragraph{ContextCite-128.}
ContextCite~\cite{r5} estimates the influence of
document units by fitting a sparse linear surrogate
to the decision model's responses under random
subsets of the document.

For an input $x=(D,q,\mathcal A)$ with document
units $u_1,\ldots,u_m$, let
$\hat a=\argmax_{a\in\mathcal A}p(a\mid x)$
denote the model's original prediction.
We sample 128 independent binary masks
$z^{(j)}\in\{0,1\}^m$, where each coordinate
is sampled independently from
$\operatorname{Bernoulli}(1/2)$.
A unit $u_i$ is retained when $z_i^{(j)}=1$
and removed otherwise. The question and answer
options remain unchanged. For each mask, we evaluate the frozen decision
model on the resulting perturbed input
$x_{z^{(j)}}$ and record the log-odds of the
original prediction:
\begin{equation}
y_j=
\log\frac{p(\hat a\mid x_{z^{(j)}})}
{1-p(\hat a\mid x_{z^{(j)}})}.
\end{equation}
We then fit a sparse linear surrogate using
$\ell_1$-regularized least squares:
\begin{equation}
(\hat\beta_0,\hat\beta)
=
\argmin_{\beta_0,\beta}
\left\{
\frac{1}{128}\sum_{j=1}^{128}
\left(y_j-\beta_0-(z^{(j)})^\top\beta\right)^2
+\lambda\|\beta\|_1
\right\},
\end{equation}
where $\beta\in\R^m$. The fitted coefficients
$\hat\beta_i$ serve as the attribution scores,
and document units are ranked by decreasing
coefficient value. 

Unlike \method{}, which trains a shared
attribution model offline, ContextCite fits a
new surrogate for every question. ContextCite-128
requires 128 perturbed model evaluations in
addition to the original decision pass, for
a total of 129 forward passes per question.

\paragraph{LIME-32.}
LIME estimates the influence of document units by
fitting a locally weighted linear surrogate to
the decision model's predictions under random
document perturbations. For an input $x=(D,q,\mathcal A)$ with $m$ document
units, let $\hat a$ denote the model's prediction
on the original input. We generate 32 binary masks
$z^{(j)}\in\{0,1\}^m$, where $z_i^{(j)}=1$ indicates
that unit $u_i$ is retained and $z_i^{(j)}=0$
indicates that it is removed. For each mask, we evaluate the decision model on
the perturbed input $x_{z^{(j)}}$ and define
\[
y_j={\ell(x_{z^{(j)}}),\qquad \ell(x')=\log\frac{p(\hat a\mid x')}{1-p(\hat a\mid x')}}.
\]
We fit a linear surrogate
$g(z)=\beta_0+z^\top\beta$, where
$\beta\in\R^m$, by minimizing the locally weighted
squared error
\begin{equation}
\min_{\beta_0,\beta}
\left\{
\sum_{j=1}^{32}
\pi_x(z^{(j)})
\left(y_j-\beta_0-(z^{(j)})^\top\beta\right)^2
+\Omega(\beta)
\right\}.
\end{equation}
Here, $\pi_x(z)$ is a proximity kernel assigning
greater weight to perturbations closer to the
original document, represented by the all-ones
mask, and $\Omega$ denotes the regularization
or feature-selection penalty. {As run, $\pi_x(z)=\sqrt{\exp(-d(z)^2/25^2)}$ with $d(z)$ one hundred times the cosine distance between $z$ and the all-ones mask, $\Omega(\beta)=\|\beta\|_2^2$ (ridge with coefficient 1) over all $m$ units, and each mask removes a number of units drawn uniformly from $1,\ldots,m$, chosen uniformly at random under a seed fixed by the question (LIME's text sampler); the fit also includes the unperturbed input, with weight 1.} The fitted coefficients $\hat\beta_i$ serve as
attribution scores, and document units are ranked
by decreasing coefficient value. LIME-32 requires
32 perturbed model evaluations per question, in
addition to the original decision pass.

\paragraph{Kernel SHAP-32.}
Kernel SHAP similarly fits a linear surrogate to
the decision model's predictions under document
perturbations, but uses a weighting kernel derived
from the Shapley-value formulation. We represent each subset of document units by
a binary mask $z\in\{0,1\}^m$ and evaluate
the model on 32 sampled subsets. As above,
the surrogate takes the form
\[
g(z)=\phi_0+z^\top\phi,
\qquad \phi\in\R^m.
\]

For a nonempty, non-full subset, let
$|z|=\sum_{i=1}^{m}z_i$ denote the number
of retained units. The SHAP kernel is
\begin{equation}
\pi_{\mathrm{SHAP}}(z)
=
\frac{m-1}
{\binom{m}{|z|}|z|(m-|z|)}.
\end{equation}
The surrogate is fitted by minimizing
the weighted squared error
\begin{equation}
\min_{\phi_0,\phi}
\sum_{j=1}^{32}
\pi_{\mathrm{SHAP}}(z^{(j)})
\left(y_j-\phi_0-(z^{(j)})^\top\phi\right)^2,
\end{equation}
subject to the endpoint constraints
\[
\phi_0={\ell(x_{\mathbf 0})},
\qquad
\phi_0+\sum_{i=1}^{m}\phi_i
={\ell(x)}.
\]
Here, $x_{\mathbf 0}$ denotes the input with
all document units removed. The resulting coefficients $\hat\phi_i$
approximate the Shapley values of the document
units. We rank units by decreasing
$\hat\phi_i$. Kernel SHAP-32 uses a budget of
32 sampled perturbations per question. {As run, the 32 subsets are drawn by SHAP's KernelExplainer sampler under a seed fixed by the question, the empty document costs one more pass, and SHAP's default feature selection (\texttt{l1\_reg="num\_features(10)"}, by least-angle regression) keeps at most 10 units before the constrained fit, so every other unit gets $\hat\phi_i=0$.}

\paragraph{{SelfElicit.}}
{SelfElicit~\cite{r20} scores units from the attention of the decision pass itself, under our decision prompt, with no extra pass and no training: the attention from $T$, averaged over the heads of each layer and summed over the unit's tokens, is divided by that layer's total over the document's units, averaged over the deeper half of the layers (on Gemma~4 and the Jeff models their global-attention layers, on the Qwen models their full-attention layers) and divided by the unit's number of tokens.}

\paragraph{{BM25.}}
{BM25~\cite{r25} ranks the units by their lexical match with the question (the criterion of the decision input, without the answer options), with $k_1=1.5$, $b=0.75$, and the inverse document frequencies and mean unit length computed over all units of the dataset; words are lowercased runs of letters and digits, and exact ties are broken by a fixed per-unit hash. It does not depend on the model.}

\paragraph{{Random.}}
{Random ranks the units by a pseudo-random key derived from the dataset, the question id and the unit index: a uniformly random permutation, the same for every model.}

\paragraph{{Attention (T).}}
{Attention (T) is the training-free score $s_i^{\mathrm{free-T}}(x)$ of \cref{sec:training_free}.}

\subsection{Converting datasets into decision questions}
\label{app:conversion-recipe}\label{app:dataset_conversion}

We convert seven datasets with human-annotated evidence into
a common decision format. Each example consists of a document
$D$, a question or decision criterion $q$, and a finite set
of answer options $\mathcal A$. The model assigns a
probability to each option and selects the most likely one.

We construct the decision questions using fixed,
dataset-specific templates. When the original dataset
already provides a set of possible answers, we retain
those choices. Otherwise, we construct suitable options.
For example, MultiRC asks whether a proposed answer is
correct, using \texttt{yes}/\texttt{no} choices. FinQA
uses the correct numerical answer and three alternative
values constructed by modifying the original calculation
or its result.

\Cref{tab:recipe} summarizes the conversion of each
dataset. We use 3,279 evaluation questions containing
at least one human-annotated evidence passage.
Questions without annotated evidence are excluded.
The correct answer and evidence annotations are retained
for evaluation but are not identified as such in the
model's input.

\begin{table}[H]
\centering
\footnotesize
\setlength{\tabcolsep}{4pt}
\renewcommand{\arraystretch}{1.12}
\begin{tabularx}{\textwidth}{@{}
  >{\raggedright\arraybackslash}p{2.5cm}
  >{\raggedright\arraybackslash}p{1.6cm}
  >{\raggedright\arraybackslash}X
  >{\raggedright\arraybackslash}p{2.0cm}
  r@{}}
\toprule
Dataset & Drawn from & Decision question and answer options
& Annotated evidence & Questions \\
\midrule

MultiRC~\cite{r16,r8}
& ERASER test set
& Is a proposed answer to the passage question correct?
  (\texttt{yes}, \texttt{no})
& Sentences
& 598 \\

SciFact~\cite{r30}
& Train and dev sets
& Does the abstract support, contradict, or provide
  insufficient information about a scientific claim?
& Rationale sentences
& 354 \\

ContractNLI~\cite{r18}
& Test set
& Does a contract entail, contradict, or not mention
  a given hypothesis?
& Evidence spans
& 206 \\

HotpotQA~\cite{r34}
& Dev set (distractor setting)
& Answer a comparison question using either
  \texttt{yes}/\texttt{no} or two compared entities.
& Supporting sentences
& 598 \\

FinQA~\cite{r4}
& Test set
& Select the correct numerical answer to a question
  about a financial report from four candidate values.
& Supporting facts
& 551 \\

Evidence Inference~\cite{lehman2019evidence,r9}
& Test set
& Determine whether a clinical intervention significantly
  increased, decreased, or did not significantly change
  an outcome relative to a comparator.
& Evidence spans
& 527 \\

ConditionalQA~\cite{r28}
& Train set
& Answer a question about government guidance for
  a specified scenario (\texttt{yes}, \texttt{no});
  only unconditionally answerable questions are used.
& HTML elements
& 445 \\

\midrule
\textbf{Total} & & & & \textbf{3,279} \\
\bottomrule
\end{tabularx}

\caption{\textbf{Conversion of seven human-annotated
datasets into decision questions.}
The column \texttt{drawn from} gives the part of each original dataset, as
released by its authors (their train, development or test set), that our
evaluation questions come from. These datasets are disjoint from \method{}'s  training and validation data (Section~\ref{sec:experimental_setup}). The \texttt{annotation column} describes the nature of the human-identified supporting evidence.
We removed questions that had no evidence annotation in their context.}
\label{tab:recipe}
\end{table}

For the FinQA-CF experiment in \cref{sec:finqa-cf},
we additionally convert 550 FinQA questions into
\texttt{yes}/\texttt{no} decisions about a proposed
numerical value. Each question is evaluated three times: with the correct value, with a wrong value, and with the placeholder \texttt{x}, while keeping the document, underlying question, and
human evidence annotations fixed. 

\subsection{Model input}
\label{app:model-input}

Each decision question is presented as two chat messages:
an instruction and a user message containing
a JSON object with three fields:
\texttt{evidence} (the document; evidence here means the document rather than the evidence we want to retrieve),
\texttt{criterion} (the decision question), and
\texttt{options} (the possible answers). 
Each answer option is assigned a letter (A, B, etc.)
and a textual description. For binary decisions, the
options are \texttt{true} and
\texttt{false}, corresponding to the
\texttt{yes}/\texttt{no} choices described above. The correct answer is included among the available
options but is not identified as correct. Human evidence
annotations are not supplied to the model either in post-training nor training \method{}.
\Cref{lst:model-input} illustrates the input format
using a MultiRC question. The document is abbreviated
for readability.

\begin{lstlisting}[
basicstyle=\ttfamily\footnotesize,
breaklines=true,
breakatwhitespace=true,
columns=fullflexible,
keepspaces=true,
showstringspaces=false,
frame=single,
caption={\textbf{Illustrative decision-model input.}
The document is abbreviated; the system instruction,
question format, and option descriptions follow the
Jeff-1.1-Large input format. The final template token
is the decision readout position $T$.},
label={lst:model-input}
]
<bos><|turn>system
Apply the supplied criterion to the supplied evidence.
Choose exactly one listed option.
Respond with only its uppercase letter, with no explanation or reasoning.<turn|>
<|turn>user
{
  "evidence": "Susan wanted to have a birthday party.\n
               ...\n
               On the day of the party, all five friends showed up.\n
               Each friend had a present for Susan.",

  "criterion": "Question about the passage: How many presents did Susan receive?\n
                Proposed answer: Five\n
                Is the proposed answer correct according to the passage?",

  "options": [
    {
      "letter": "A",
      "description": "true: The passage shows that the proposed answer is correct."
    },
    {
      "letter": "B",
      "description": "false: According to the passage, the proposed answer is not correct."
    }
  ]
}<turn|>
<|turn>model
<|channel>thought
<channel|>
\end{lstlisting}

The model's chat template adds the special tokens such as  \texttt{<bos>}, \texttt{<turn|>}, \texttt{<|channel>}, etc., with intermediate reasoning generation
disabled. No answer or reasoning tokens are generated and 
the decision probabilities are computed directly at
the final prompt position $T$. The Qwen-based models
use the same message structure with different chat
template tokens.

\subsection{Context segmentation, evidence alignment, data statistics}
\label{app:segmentation}
This section explains
how we segment context into units, how we decide which units
count as human-annotated evidence, and how closely our units
match the original annotations. We provide some concrete examples of segmentation and statistics on how our units line up with the human-annotated data.

Recall that every method in this paper aims to rank the
parts of context by how likely they are to be the evidence
for the question. To compare methods we must first fix what a
``part'' is. We call these parts (context) \emph{units}. 

\begin{figure}[!htbp]
\centering
\includegraphics[width=\linewidth]{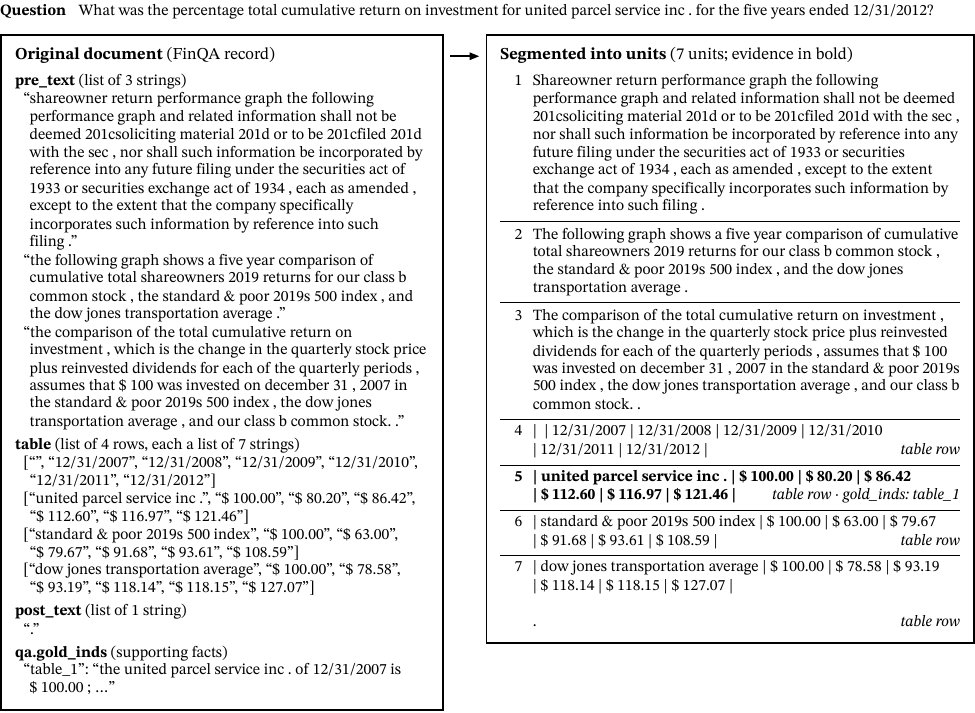}
\caption{{\textbf{FinQA example} Left: FinQA raw document data stores three fields \texttt{pre\_text, table, post\_text}. \texttt{pre\_text} denotes the list of the sentences before the table , \texttt{table} denotes the table as a list of rows of cells and \texttt{post\_text} denotes  the list of the sentences after it. Right: Our segmentation method partitions the text into seven units. We capitalize the first letter of each sentence and join the cells of a row with vertical bars. The text after the table is a single period with no word in it, so it joins unit 7. The FinQA annotators marked the United Parcel Service row as the supporting fact, so unit 5 (bold) is the evidence unit for this question.}}
\label{fig:unit-example}
\end{figure}

\begin{figure}[!htbp]
\centering
\includegraphics[width=\linewidth]{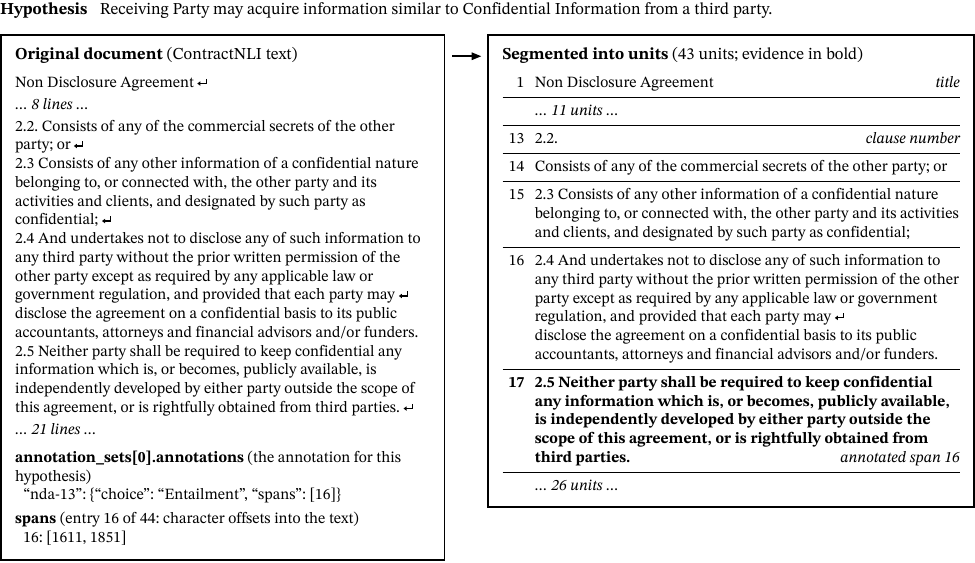}
\caption{\textbf{Example of segmenting an ContractNLI document.} Left: the title, clauses 2.2 to 2.5, and annotation information of a ContractNLI agreement are shown as raw text with other context abbreviated for brevity. A small return arrow marks each line break of the original, in both boxes. Right: the same text cut into units by our segmentation method and numbered in document order. The two entries at the foot of the left box are ContractNLI's own evidence annotation. In particular for \texttt{nda-13} the annotators chose Entailment and pointed to span 16 of the contract. Span 16 is characters 1611 to 1851 of the text, which is clause 2.5. That span lies our unit 17, so unit 17 so we mark this as the evidence unit (show in bold). 
}
\label{fig:unit-example-text}
\end{figure}

\paragraph{Context units.}
Our segmented units are usually one sentence, one table row, one object,
or one heading. Figure~\ref{fig:unit-example} shows a short FinQA
document cut into seven units: three sentences and four table
rows. Figure~\ref{fig:unit-example-text} shows the same for an excerpt of a ContractNLI agreement. We note that using one segmentation method may not cover the intricacies of all datasets. For example, in Figure~\ref{fig:unit-example-text} shows our segmentation treats the clause "2.2." as its own unit (unit 13) which arguably should be merged with unit 14. In general, one address these issues by designing a segmentation method per task-relevant documents. Finally, we provide statistics of the units per dataset in Table~\ref{tab:unit-coverage}.

\paragraph{Context segmentation.}
Five of the seven datasets (MultiRC, SciFact, HotpotQA, FinQ, ConditionalQA)
contain documents that are pre-divided into units. These are provided as \texttt{python List} where elements in the list are their units and a subset of which are annotated as ground-truth evidence. Our loader writes each of the original units as its own line, and our line-break rule treats each as its own unit. Hence their pre-divided units and our generated units from their dataset lines up.

For the other datasets, we follow roughly three rules:
\begin{itemize}[leftmargin=*,itemsep=2pt,topsep=3pt]
\item \emph{Structure first.} 
We leave all document text as it is (i.e. no deletion). FinQA's text is lower-cased, so we also capitalize the first letter of each of its sentences.  We place the section headers of Evidence Inference on their own lines. 
\item \emph{Line breaks.} A line break starts a new unit,
unless the next line appears to continue a wrapped sentence.
\item \emph{Sentence ends.} Within a line, a unit ends at a
period, question mark, or exclamation mark that is followed
by a space and a capital letter or digit. To mitigate early segmentation we enforce rules that if a period ends after a set of patterns then we continue the segment (for e.g., after a single-letter initial).
\end{itemize}
The rules read only the document text. They never see the
question, the answer options, or the evidence annotations, and
they run before any model does. Every model and every
attribution method therefore ranks exactly the same units.
Using sentences and lines as units follows
ContextCite~\cite{r5}.

\paragraph{Capping context to 64 units.}
We cap the number of units at 64 to bound the cost of leave-one-out, which needs on forward pass per unit. The choice was made prior to seeing any performance results on our evaluation datasets. When a document has more than 64 units, we repeatedly merge the two neighbouring
units with the shortest combined length until 64 remain. No
text is ever dropped and long documents simply end up with
coarser units. The cap affects 689 of the 3,279 questions (21\%). Of these, 644 are in three datasets: 105 in ContractNLI, 203 in Evidence Inference, and 336 in ConditionalQA.

\paragraph{Effects of capping.} We find no sign that the choice of 64 units favours Jevidence over competing baselines. The 64 units is a cap that is enforced to each document alone, and all baseline methods run on these capped documents. We ran a few tests: (i) We reran Jevidence and several baslines with no merging (i.e. every unit of each document was kept) and the retrieval performance gap between Jevidence and the baselines remained within $1-2$\% difference and (ii) compared performance of Jevidence on merged vs unmerged questions: we found Jevidence's performance on Hit@1 and AUROC dropped by $\sim$1\%, and Recall@5 dropped about 3-5\%.

\paragraph{How we mark our units as evidence.}
The source datasets mark their evidence in different forms:
sentences, table rows, document elements, or character spans.
We map all of them onto our units with one rule. A unit is
\emph{marked} as evidence if it overlaps an annotated span by
at least half the length, in characters, of the shorter of the
two. Every other unit is unmarked (Section~\ref{sec:experimental_setup}). Note that the original document is kept and no text is deleted.

\paragraph{Coverage statistics of evidence.}
We use two terms. An \emph{annotated span} is a piece of the original text that the dataset's annotators
identified as evidence. A \emph{marked unit} is one of our units that we label as evidence because it overlaps
an annotated span, by the rule above. Table~\ref{tab:unit-coverage} compares the two in both directions. In one
direction, the annotators' evidence is kept: every annotated span has at least one marked unit, and 99.9\% of
the text in annotated spans lies inside a marked unit. In the other, marked units contain little else: only
2.9\% of them are mostly other text, meaning that less than half of their characters lie inside an annotated
span.

\begin{table}[!htbp]
\centering
\footnotesize
\setlength{\tabcolsep}{4pt}
\renewcommand{\arraystretch}{1.1}
\newcolumntype{R}[1]{>{\raggedleft\arraybackslash}p{#1}}
\begin{tabular}{@{}l r R{1.35cm} R{1.2cm} R{1.45cm} R{2.0cm} R{2.0cm}@{}}
\toprule
& & \multicolumn{2}{c}{Unit size (median)}
  & \multicolumn{3}{c}{Evidence labels} \\
\cmidrule(lr){3-4}\cmidrule(l){5-7}
Dataset & Questions
& Units per document
& Words per unit
& Marked units per question
& Annotated text inside marked units
& Marked units that are mostly other text \\
\midrule
MultiRC            & 598 & 14 & 15 & 2.3 & 100.0\% & 0.0\% \\
SciFact            & 354 & 9  & 22 & 1.8 & 100.0\% & 0.0\% \\
ContractNLI        & 206 & 64 & 24 & 1.8 & 100.0\% & 10.8\% \\
HotpotQA           & 598 & 48 & 17 & 2.4 & 99.9\%  & 0.3\% \\
FinQA              & 551 & 30 & 21 & 1.8 & 100.0\% & 0.0\% \\
Evidence Inference & 527 & 51 & 22 & 2.5 & 100.0\% & 2.1\% \\
ConditionalQA      & 445 & 64 & 19 & 3.7 & 99.4\%  & 9.2\% \\
\midrule
\textbf{All}       & \textbf{3,279} & \textbf{36} & \textbf{19} & \textbf{2.4}
                   & \textbf{99.9\%} & \textbf{2.9\%} \\
\bottomrule
\end{tabular}

\caption{\textbf{Coverage of human-annotated evidence.}
The first two columns give the typical size of a document and
of a unit. The last three describe the evidence labels.
\emph{Marked units per question} is the average number of
units labelled as evidence.
\emph{Annotated text inside marked units} is the share of the
annotators' evidence that our marked units contain (100\%
means that no evidence is lost).
\emph{Marked units that are mostly other text} is the share
of marked units in which less than half of the text was
annotated (0\% means that every marked unit is mostly
evidence). These statistics are computed after capping at 64 units.} 
\label{tab:unit-coverage}
\end{table}

{Table~\ref{tab:input-size} compares the size of the training and evaluation inputs. The training
questions are much shorter: a median of 4 units and 197 tokens, against 36 units and 1,190 tokens
for the evaluation questions.}

\begin{table}[H]
\centering
\footnotesize
\setlength{\tabcolsep}{5pt}
\begin{tabular}{@{}l r rrr rrr@{}}
\toprule
& & \multicolumn{3}{c}{Units per document} & \multicolumn{3}{c}{Tokens per input} \\
\cmidrule(lr){3-5}\cmidrule(l){6-8}
Dataset & Questions & Median & Mean & Max & Median & Mean & Max \\
\midrule
MultiRC            & 598 & 14 & 14.2 & 23 & 459 & 465 & 730 \\
SciFact            & 354 & 9  & 10.1 & 25 & 460 & 535 & 1,583 \\
ContractNLI        & 206 & 64 & 56.1 & 64 & 2,025 & 2,493 & 8,101 \\
HotpotQA           & 598 & 48 & 47.6 & 64 & 1,388 & 1,391 & 2,825 \\
FinQA              & 551 & 30 & 29.3 & 64 & 1,159 & 1,178 & 2,432 \\
Evidence Inference & 527 & 51 & 48.6 & 64 & 2,221 & 2,653 & 8,186 \\
ConditionalQA      & 445 & 64 & 60.5 & 64 & 1,713 & 1,969 & 7,647 \\
\midrule
\textbf{All evaluation} & \textbf{3,279} & \textbf{36} & \textbf{36.8} & \textbf{64} & \textbf{1,190} & \textbf{1,444} & \textbf{8,186} \\
\midrule
\textbf{Training} & \textbf{44,747} & \textbf{4} & \textbf{8.5} & \textbf{64} & \textbf{197} & \textbf{299} & \textbf{4,038} \\
\bottomrule
\end{tabular}
\caption{{\textbf{Size of training and evaluation inputs.} Units per document and tokens per input
(the document together with the question and the answer options), for the marked questions of each
evaluation dataset and for the 44,747 training questions. Evaluation unit counts are after merging
to at most 64 units. Token counts here use the tokenizer of the Gemma-based models.}}
\label{tab:input-size}
\end{table}

\begin{figure}[H]
\centering
\includegraphics[width=\linewidth]{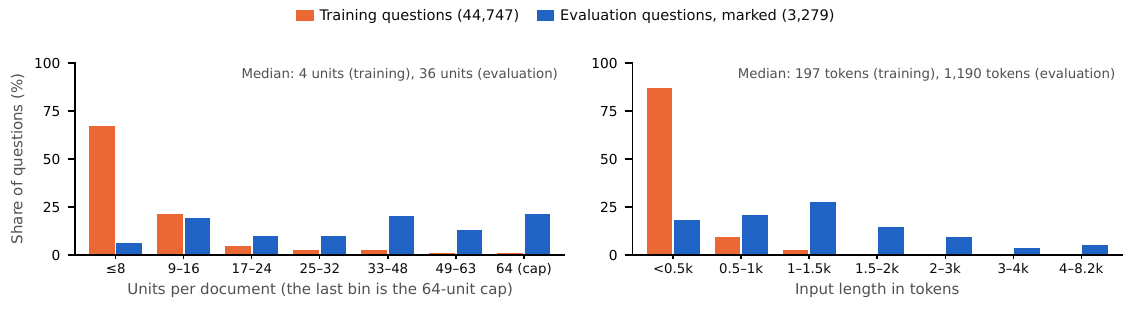}
\caption{{\textbf{Distribution of training and evaluation input sizes.} Share of questions (\%) per bin for the 44,747 training questions (orange) and the 3,279 marked evaluation questions (blue). Left: units per document  which are counted after merging to at most 64, and the last bin is the 64-unit cap. Right: input length in tokens, for the full input (document, question and answer options) with the tokenizer of the Gemma-based models. The bins are those of Figure~\ref{fig:length}. Per-dataset statistics are in Table~\ref{tab:input-size}.}}
\label{fig:length-dist}
\end{figure}

{Figure~\ref{fig:length-dist} shows that \method{} training dataset has much shorter context than our evaluation dataset. In particular, the two distributions in the bins used in Figure~\ref{fig:length}: 67\% of the training questions have at most 8 units and 87\% have fewer than 500 tokens, against 6\% and 18\% of the evaluation questions.}

\clearpage
\section{Additional results}\label{app:additional-results}

\subsection{Ablation comparing AT2 to \method{}}\label{app:at2-to-jevidence}
\begin{figure}[!htbp]
\centering
\includegraphics[width=\linewidth]{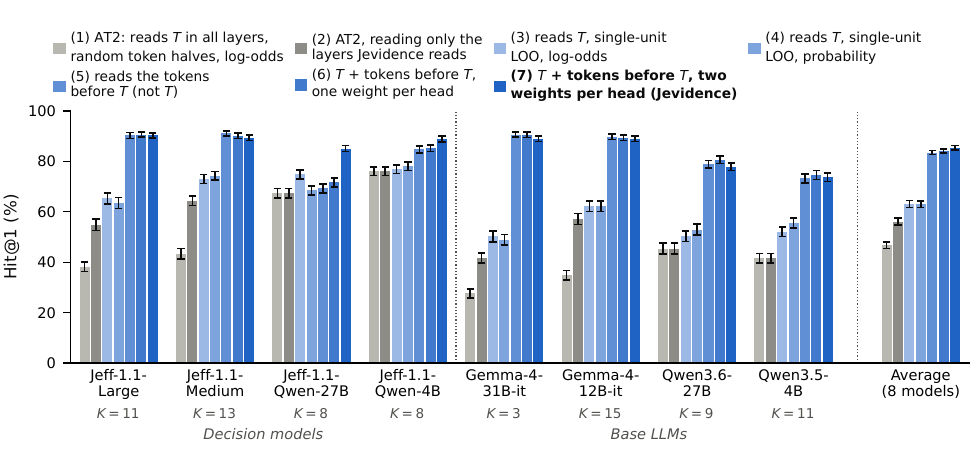}
\caption{\textbf{From AT2 to \method{}.} Hit@1 over the 3,279 marked questions for each of the eight models and their
average. Bars (1) to (7) are the steps listed in the text, each changing one thing from the bar before it. $K$ under each
model is the window of \method{}, selected on development questions. Error bars are 95\% bootstrap intervals over
documents.}
\label{fig:at2-ladder}
\end{figure}

In \Cref{fig:at2-ladder}, we show the effect of each of the changes we made to AT2 to turn it into \method{}. Each bar changes one thing from the bar before it, and bars (5) to (7) each use their own window $K$, chosen held-out questions. The value of $K$ depicted in the figure corresponds to the one used for \method{}, i.e. the right-most bar for each model. The variants included in the plot are:
\begin{enumerate}[label=(\arabic*),nosep,leftmargin=2.4em]
\item AT2, as we adapt it to decision models, learns one weight per head on the attention from the readout position $T$
in every softmax-attention layer, trained with log-odds targets and random masking of half the tokens in the document. Average Hit@1: 46.8\%.
\item AT2, but only reading the global attention layers, excluding sliding-window attention layers (which is the same set of layers that \method{} reads). On Qwen models, this is the same as (1). Average Hit@1: 56.0\%.
\item Replace randomly masking half the tokens with masking out one unit at a time; the targets become the corresponding LOO drops in log-odds. Average Hit@1: 63.1\%.
\item Replaces the drop in log-odds with the drop in probability. Average Hit@1: 63.1\%.
\item The probe reads the mean attention over the $K$ positions before $T$ instead of the attention at $T$. $K$ is fitted on held-out data. Average Hit@1: 83.5\%.
\item The probe reads the mean over $T$ and the $K$ positions before it, with one weight per head. Average Hit@1: 84.1\%.
\item The probe reads the attention at $T$ and the mean over the $K$ positions before $T$ as separate features, with two
weights per head, which is \method{}. Average Hit@1: 85.4\%.
\end{enumerate}

These results show that the largest gain stems from reading out attention at positions before $T$ (change (5)); surprisingly, adding $T$ into the readout only slightly improves Hit@1.\\ We employed drop in probability (change (4)) as an easier objective more aligned with LOO, but this ablation shows that it does not lead to an improvement on downstream Hit@1.\\ 
A separate set of weights for position $T$ (change (7)) mostly improves Qwen models, and actually slightly reduces Hit@1 on Gemma models. We kept the same settings across all models for simplicity, but these results show that model-tuned settings could improve downstream performance. 

\FloatBarrier

\paragraph{Effect of attention positions.}
\Cref{tab:design} isolates one design choice of \method{}: the prompt positions from which the probe takes
attention. Recall that $A_t(x)\in\R^{m\times H}$ holds the attention from prompt position $t$ to each of the
$m$ units, for each of the $H$ heads (\cref{eq:features}), and that $T$ is the final prompt token. We compare
three feature matrices:
\begin{align*}
B_{\mathrm{readout}}(x) &:= A_T(x)\in\R^{m\times H} && \text{(readout only)},\\
B_{\mathrm{preceding}}(x) &:= \frac{1}{K}\sum_{k=1}^{K}A_{T-k}(x)\in\R^{m\times H} && \text{(preceding only)},\\
B(x) &:= \bigl[\,B_{\mathrm{readout}}(x)\;\big|\;B_{\mathrm{preceding}}(x)\,\bigr]\in\R^{m\times 2H} && \text{(readout + preceding)}.
\end{align*}
The three probes are trained on the same questions, with the same leave-one-out targets and loss.

Table~\ref{tab:design} shows the token positions before $T$ carry most of the signal. Going from $B_{\mathrm{preceding}}$ to $B$, that is, adding the attention from $T$ changes little on the Gemma models and helps on the two Qwen-based post-trained models (from 69.3\% to 85.1\% on Jeff-1.1-Qwen-27B and from 84.8\% to 88.8\% on Jeff-1.1-Qwen-4B).

\begin{table}[H]\centering\small\setlength{\tabcolsep}{5pt}
{
\begin{tabular}{l@{\hspace{14pt}}c@{\hspace{18pt}}cc@{\hspace{18pt}}cc}\toprule
 & \multicolumn{1}{c}{Readout only} & \multicolumn{2}{c}{Preceding only} & \multicolumn{2}{c}{Readout + preceding (\method{})} \\
 & \multicolumn{1}{c}{$B_{\mathrm{readout}}(x)$} & \multicolumn{2}{c}{$B_{\mathrm{preceding}}(x)$} & \multicolumn{2}{c}{$B(x)$} \\
\cmidrule(lr){2-2}\cmidrule(lr){3-4}\cmidrule(lr){5-6}
Model & Hit@1 (\%) & Hit@1 (\%) & $K$ & Hit@1 (\%) & $K$\\\midrule
\multicolumn{6}{l}{\emph{Post-trained models}}\\
Jeff-1.1-Large & 63.5 & 90.3 & 9 & 90.3 & 11\\
Jeff-1.1-Medium & 74.4 & 91.1 & 5 & 89.4 & 13\\
Jeff-1.1-Qwen-27B & 68.5 & 69.3 & 5 & 85.1 & 8\\
Jeff-1.1-Qwen-4B & 78.0 & 84.8 & 5 & 88.8 & 8\\
\addlinespace[3pt]
\multicolumn{6}{l}{\emph{Base LLMs}}\\
Gemma-4-31B-it & 48.9 & 90.6 & 11 & 88.9 & 3\\
Gemma-4-12B-it & 62.2 & 89.9 & 8 & 89.0 & 15\\
Qwen3.6-27B & 52.9 & 78.8 & 8 & 77.9 & 9\\
Qwen3.5-4B & 55.5 & 73.3 & 8 & 73.8 & 11\\
\bottomrule\end{tabular}
}
\caption{\textbf{Concentration of evidence signal.} 
Ablation of  three linear probes whose only difference is where they read their attention features. \emph{Readout only} ($B_{\mathrm{readout}}$): the attention from the final prompt token $T$, the position at which the decision is read. \emph{Preceding only} ($B_{\mathrm{preceding}}$): the attention averaged over the $K$ tokens immediately before $T$. \emph{Readout + preceding} ($B$): both, as separate features (\method{}). $K$ is selected separately for each model and each probe on the held-out validation questions (Appendix~\ref{app:reader}). Hit@1 is the percentage of the 3,279 marked questions for which the top-ranked unit is human-marked.}
\label{tab:design}
\end{table}

\subsection{Expanded main results}\label{app:performance-results}

This section shows accuracy of the Jeff-1.1 models across our validation datasets, as well as several detailed retrieval results split by datasets.

\paragraph{Jeff-1.1 accuracy.} Every Jeff-1.1 model answers more of the marked questions correctly than its base
LLM (89.3--94.2\% against 78.0--91.1\%). Most of this gain comes from FinQA, where the base LLMs answer 34.1--85.3\% of
the questions correctly and the post-trained models 84.8--98.5\%. ContractNLI is the one dataset where three of the four
post-trained models score below their base LLM.

\paragraph{Retrieval across methods.} Table~\ref{tab:all-methods-with-sum} reports Hit@1, Recall@5, and AUROC of
every method on the eight models. \method{} has the highest Hit@1 on six of them. ContextCite-128 is higher on
Jeff-1.1-Qwen-27B (87.1 against 85.1) and on Qwen3.6-27B (81.2 against 77.9). The training-free sum reaches 76.4--86.5\%
Hit@1 on the Gemma~4 models, against 88.9--90.3\% for \method{}, but only 12.6--45.3\% on the Qwen models.

\paragraph{Retrieval by dataset.} Tables~\ref{tab:all-methods-sets} and~\ref{tab:all-methods-sets-pp} show
Hit@1 by dataset and across all methods and Jeff models and their backbones. On the post-trained models, \method{} is best on 27 of the 28 pairs of model and dataset. The exception is
FinQA on Jeff-1.1-Qwen-27B (69.3\% against 96.6\% for ContextCite-128), which alone accounts for ContextCite-128's lead on
that model. We note that \method{} varies a lot with $k$ on this dataset, and with $k=7$ instead of $k=8$, it achieves 93.5 instead of 69.3. On the base LLMs, \method{} is best on all seven datasets for Gemma-4-31B-it and on six for Gemma-4-12B-it,
but only on three for each of the two Qwen models, where ContextCite-128 is best on most of the others. Recall that ContextCite-128 requires 128 additional forward passes through the model, so even on datasets where it is ahead of \method{} in the metrics of these tables, it remains prohibitively expensive to run in practice.

\begin{figure}[!htbp]
\centering
\includegraphics[width=\linewidth]{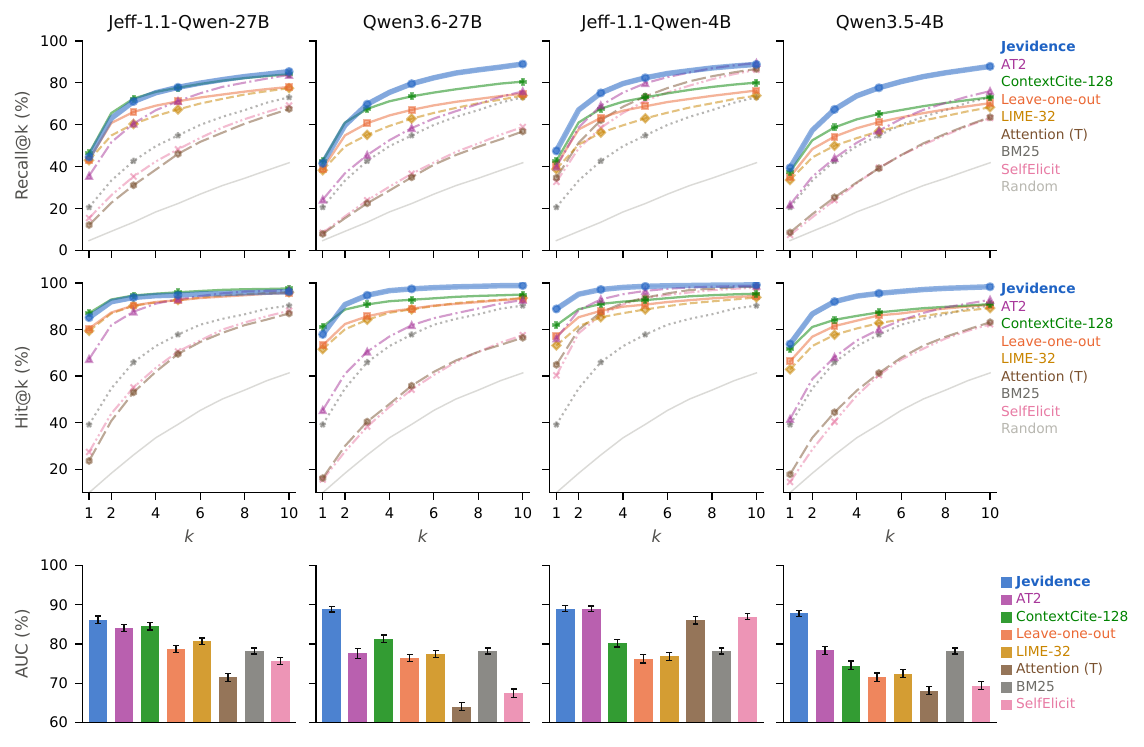}
\caption{{\textbf{Retrieval quality on Qwen post-trained models and their backbones.} The Random method randomly selects a subset of $k$ units. Black error bars are 95\% confidence intervals. This figure shows Qwen models; Figure~\ref{fig:recall-k} has the same plot for Gemma~4 models.}}
\label{fig:recall-k-qwen}
\end{figure}

\paragraph{Qwen models.} Figure~\ref{fig:recall-k-qwen} repeats Figure~\ref{fig:recall-k} for the Qwen models.

\paragraph{Training-free attribution.} Table~\ref{tab:all-methods-with-sum} compares the training-free sum with \method{} and the baselines which shows competitive performance with \method{} for Gemma models but vastly underperformance on Qwen models. This difference motivates future work to explore whether selecting heads and positions for each model can make the training-free rule more reliable. 

\begin{table}[H]\centering\footnotesize\setlength{\tabcolsep}{3.6pt}
\begin{tabular}{lrrrrrrr@{\hspace{12pt}}r}\toprule
 & MultiRC & SciFact & ContractNLI & HotpotQA & FinQA & Evidence Inf. & ConditionalQA & All\\
Model & (598) & (354) & (206) & (598) & (551) & (527) & (445) & (3,279)\\\midrule
\multicolumn{9}{l}{\emph{Post-trained models}}\\
\quad Jeff-1.1-Large & 93.0 & 87.6 & 86.9 & 97.2 & 97.5 & 92.8 & 95.7 & 93.9\\
\quad Jeff-1.1-Medium & 93.1 & 84.7 & 85.0 & 95.3 & 93.5 & 91.1 & 90.8 & 91.5\\
\quad Jeff-1.1-Qwen-27B & 93.0 & 87.6 & 85.0 & 97.8 & 98.5 & 94.5 & 94.8 & 94.2\\
\quad Jeff-1.1-Qwen-4B & 91.3 & 82.5 & 81.6 & 94.5 & 84.8 & 91.7 & 91.2 & 89.3\\
\addlinespace[3pt]
\multicolumn{9}{l}{\emph{Base LLMs}}\\
\quad Gemma-4-31B-it & 92.5 & 86.4 & 90.8 & 95.2 & 85.3 & 91.8 & 93.7 & 91.1\\
\quad Gemma-4-12B-it & 93.1 & 84.2 & 85.4 & 93.0 & 43.4 & 89.9 & 90.1 & 82.4\\
\quad Qwen3.6-27B & 91.6 & 86.2 & 87.4 & 95.3 & 57.5 & 92.4 & 93.5 & 86.1\\
\quad Qwen3.5-4B & 88.1 & 89.3 & 80.6 & 87.1 & 34.1 & 86.9 & 85.6 & 78.0\\
\bottomrule\end{tabular}
\caption{\textbf{Decision accuracy of each model.} The percentage of the marked questions that the model answers correctly.}
\label{tab:decision-accuracy}
\end{table}

\begin{table}[H]\centering\footnotesize\setlength{\tabcolsep}{3pt}
\begin{tabular}{lrrrrrrrrrrrr}\toprule
\emph{Post-trained models} & \multicolumn{3}{c}{Jeff-1.1-Large} & \multicolumn{3}{c}{Jeff-1.1-Medium} & \multicolumn{3}{c}{Jeff-1.1-Qwen-27B} & \multicolumn{3}{c}{Jeff-1.1-Qwen-4B}\\
\cmidrule(lr){2-4}\cmidrule(lr){5-7}\cmidrule(lr){8-10}\cmidrule(lr){11-13}
Method & Hit@1 & R@5 & AUROC & Hit@1 & R@5 & AUROC & Hit@1 & R@5 & AUROC & Hit@1 & R@5 & AUROC\\
\midrule
\method{} (ours) & \textbf{90.3} & \textbf{85.0} & \textbf{92.5} & \textbf{89.4} & {\textbf{85.4}} & 92.5 & 85.1 & \textbf{77.9} & \textbf{86.2} & \textbf{88.8} & \textbf{82.4} & \textbf{89.0}\\
Training-free sum (ours) & {76.4} & {77.8} & {90.0} & {86.5} & {84.7} & {\textbf{92.8}} & {12.6} & {30.9} & {62.7} & {45.3} & {62.6} & {80.8}\\

\noalign{\vskip 1pt}\multispan{13}{\color{black!60}\dotfill}\cr\noalign{\vskip 1pt}
{Attention (T)} & {15.6} & {31.4} & {62.9} & {21.7} & {38.9} & {69.4} & {23.5} & {46.1} & {71.5} & {64.9} & {73.5} & {86.0}\\
AT2 & 54.9 & 63.2 & 81.1 & 64.4 & 72.8 & 86.4 & 67.3 & 71.3 & 84.1 & 76.2 & 79.8 & \textbf{89.0}\\
Leave-one-out & 78.4 & 68.3 & 77.8 & 75.0 & 64.9 & 75.8 & 80.4 & 71.2 & 78.7 & 77.1 & 68.8 & 76.2\\
ContextCite-128 & 86.4 & 77.5 & 84.5 & 82.8 & 74.1 & 81.4 & \textbf{87.1} & 77.7 & 84.5 & 81.9 & 73.1 & 80.2\\
LIME-32 & 78.9 & 67.6 & 80.6 & 74.1 & 64.2 & 77.8 & 79.4 & 67.3 & 80.6 & 73.2 & 63.0 & 76.8\\
Kernel SHAP-32 & 62.6 & 58.3 & 73.8 & 59.1 & 55.0 & 71.2 & 62.2 & 57.3 & 72.8 & 57.3 & 53.3 & 69.8\\
SelfElicit & 10.6 & 29.7 & 64.1 & 16.6 & 37.7 & 73.1 & 27.4 & 48.3 & 75.6 & 60.3 & 71.5 & 87.0\\
BM25 & 39.2 & 54.9 & 78.2 & 39.2 & 54.9 & 78.2 & 39.2 & 54.9 & 78.2 & 39.2 & 54.9 & 78.2\\
Random & 9.9 & 22.4 & 50.8 & 9.9 & 22.4 & 50.8 & 9.9 & 22.4 & 50.8 & 9.9 & 22.4 & 50.8\\
\midrule
\emph{Base LLMs} & \multicolumn{3}{c}{Gemma-4-31B-it} & \multicolumn{3}{c}{Gemma-4-12B-it} & \multicolumn{3}{c}{Qwen3.6-27B} & \multicolumn{3}{c}{Qwen3.5-4B}\\
\cmidrule(lr){2-4}\cmidrule(lr){5-7}\cmidrule(lr){8-10}\cmidrule(lr){11-13}
Method & Hit@1 & R@5 & AUROC & Hit@1 & R@5 & AUROC & Hit@1 & R@5 & AUROC & Hit@1 & R@5 & AUROC\\
\midrule
\method{} (ours) & \textbf{88.9} & \textbf{84.5} & {\textbf{91.7}} & \textbf{89.0} & 82.3 & 88.9 & 77.9 & \textbf{79.6} & \textbf{88.8} & \textbf{73.8} & \textbf{77.7} & \textbf{87.8}\\
Training-free sum (ours) & {80.9} & {80.6} & {91.1} & {83.1} & {\textbf{83.8}} & {\textbf{92.4}} & {13.5} & {29.1} & {60.1} & {14.7} & {33.0} & {64.2}\\

\noalign{\vskip 1pt}\multispan{13}{\color{black!60}\dotfill}\cr\noalign{\vskip 1pt}
{Attention (T)} & {16.6} & {30.7} & {61.0} & {19.7} & {37.4} & {67.1} & {16.2} & {34.9} & {64.0} & {17.8} & {39.2} & {68.1}\\
AT2 & 41.5 & 55.5 & 75.5 & 57.1 & 67.5 & 83.0 & 45.3 & 58.4 & 77.6 & 41.6 & 57.2 & 78.4\\
Leave-one-out & 62.9 & 56.8 & 72.0 & 63.9 & 58.8 & 71.5 & 73.2 & 67.1 & 76.4 & 66.3 & 61.3 & 71.5\\
ContextCite-128 & 79.2 & 72.1 & 81.4 & 74.6 & 66.7 & 76.0 & \textbf{81.2} & 73.5 & 81.3 & 71.7 & 65.1 & 74.5\\
LIME-32 & 66.2 & 60.7 & 77.2 & 64.1 & 57.3 & 73.1 & 71.6 & 62.9 & 77.4 & 62.9 & 56.9 & 72.4\\
Kernel SHAP-32 & 54.4 & 53.7 & 70.6 & 46.4 & 46.2 & 65.0 & 54.0 & 52.3 & 69.9 & 41.5 & 43.4 & 63.6\\
SelfElicit & 11.0 & 26.7 & 60.5 & 11.4 & 31.6 & 66.2 & 15.6 & 36.6 & 67.4 & 14.5 & 39.4 & 69.4\\
BM25 & 39.2 & 54.9 & 78.2 & 39.2 & 54.9 & 78.2 & 39.2 & 54.9 & 78.2 & 39.2 & 54.9 & 78.2\\
Random & 9.9 & 22.4 & 50.8 & 9.9 & 22.4 & 50.8 & 9.9 & 22.4 & 50.8 & 9.9 & 22.4 & 50.8\\
\bottomrule\end{tabular}
\caption{\textbf{Per-dataset breakdown of post-trained Gemma/Qwen models.} Results pooled over the 3,279 questions across seven datasets.  Context (ignoring question, and answer options) contain a median of 36 units (range 3--64). Random ranking yields 9.9\% Hit@1. \method{} scores average three training seeds. Bold marks the best score per column, including rounded ties. The training-free sum averages over $T$, $T-1$, $T-2$ and $T-3$ the attention summed only over the global-attention layers.  Attention (T) uses the attention from the last prompt token $T$ alone, summed over each unit's tokens and over only the global-attention layers.}
\label{tab:all-methods-with-sum}
\end{table}
\FloatBarrier
\begin{table}[H]\centering\footnotesize\setlength{\tabcolsep}{3.6pt}\renewcommand{\arraystretch}{0.92}
{
\begin{tabular}{lrrrrrrr@{\hspace{12pt}}r}\toprule
\emph{Post-trained models} & MultiRC & SciFact & ContractNLI & HotpotQA & FinQA & Evidence Inf. & ConditionalQA & All\\
Method & (598) & (354) & (206) & (598) & (551) & (527) & (445) & (3,279)\\\midrule
\multicolumn{9}{l}{\textit{Jeff-1.1-Large}}\\
\quad \method{} (ours) & \textbf{84.4} & \textbf{81.4} & \textbf{91.7} & \textbf{89.5} & \textbf{96.9} & \textbf{94.9} & \textbf{92.1} & \textbf{90.3}\\
\quad Training-free sum (ours) & {75.1} & {74.3} & {86.4} & {75.8} & {69.1} & {76.9} & {84.3} & {76.4}\\

\noalign{\vskip 1pt}\multispan{9}{\color{black!60}\dotfill}\cr\noalign{\vskip 1pt}
\quad {Attention (T)} & {17.2} & {31.6} & {11.7} & {7.9} & {11.6} & {21.8} & {10.6} & {15.6}\\
\quad AT2 & 27.9 & 60.5 & 50.0 & 42.0 & 69.5 & 80.6 & 57.5 & 54.9\\
\quad Leave-one-out & 77.8 & 63.8 & 84.5 & 82.8 & 90.2 & 69.3 & 78.2 & 78.4\\
\quad ContextCite-128 & 78.6 & 70.9 & 86.9 & 88.5 & 93.8 & 92.6 & 89.7 & 86.4\\
\quad LIME-32 & 76.6 & 68.4 & 86.4 & 78.4 & 88.0 & 78.0 & 77.5 & 78.9\\
\quad Kernel SHAP-32 & 71.4 & 70.3 & 64.6 & 49.8 & 77.1 & 56.0 & 50.6 & 62.6\\
\quad SelfElicit & 13.0 & 33.6 & 2.4 & 3.8 & 11.1 & 8.2 & 4.3 & 10.6\\
\addlinespace[3pt]
\multicolumn{9}{l}{\textit{Jeff-1.1-Medium}}\\
\quad \method{} (ours) & \textbf{84.3} & \textbf{82.2} & \textbf{87.9} & \textbf{93.6} & \textbf{91.8} & \textbf{93.2} & \textbf{89.7} & \textbf{89.4}\\
\quad Training-free sum (ours) & {83.4} & {78.2} & {84.0} & {89.8} & {88.0} & {90.5} & {87.0} & {86.5}\\
\noalign{\vskip 1pt}\multispan{9}{\color{black!60}\dotfill}\cr\noalign{\vskip 1pt}
\quad {Attention (T)} & {21.9} & {35.9} & {18.0} & {13.5} & {15.6} & {32.4} & {17.3} & {21.7}\\
\quad AT2 & 55.0 & 75.4 & 61.7 & 65.2 & 55.9 & 83.3 & 56.6 & 64.4\\
\quad Leave-one-out & 71.1 & 68.6 & 78.2 & 78.4 & 83.3 & 71.2 & 73.3 & 75.0\\
\quad ContextCite-128 & 71.1 & 74.6 & 85.0 & 86.1 & 88.6 & 89.6 & 84.7 & 82.8\\
\quad LIME-32 & 68.9 & 69.2 & 79.6 & 74.1 & 82.2 & 76.9 & 69.0 & 74.1\\
\quad Kernel SHAP-32 & 63.9 & 68.9 & 63.6 & 50.3 & 71.7 & 53.1 & 46.1 & 59.1\\
\quad SelfElicit & 29.9 & 40.7 & 5.8 & 4.2 & 14.2 & 10.2 & 11.5 & 16.6\\
\addlinespace[3pt]
\multicolumn{9}{l}{\textit{Jeff-1.1-Qwen-27B}}\\
\quad \method{} (ours) & \textbf{82.4} & \textbf{84.2} & \textbf{86.9} & \textbf{88.3} & 69.3 & \textbf{94.9} & \textbf{92.4} & 85.1\\
\quad Training-free sum (ours) & {24.6} & {8.8} & {2.4} & {15.2} & {8.2} & {9.7} & {9.4} & {12.6}\\
\noalign{\vskip 1pt}\multispan{9}{\color{black!60}\dotfill}\cr\noalign{\vskip 1pt}
\quad {Attention (T)} & {25.4} & {23.2} & {3.9} & {25.6} & {28.9} & {31.9} & {11.2} & {23.5}\\
\quad AT2 & 55.9 & 65.8 & 63.1 & 63.5 & 67.7 & 89.4 & 64.3 & 67.3\\
\quad Leave-one-out & 78.3 & 64.1 & 82.5 & 86.3 & 90.9 & 73.8 & 82.2 & 80.4\\
\quad ContextCite-128 & 80.9 & 72.0 & 82.5 & 87.5 & \textbf{96.6} & 93.0 & 90.1 & \textbf{87.1}\\
\quad LIME-32 & 77.9 & 68.9 & 82.0 & 77.4 & 90.6 & 78.7 & 78.2 & 79.4\\
\quad Kernel SHAP-32 & 69.6 & 71.2 & 63.1 & 48.8 & 79.9 & 56.9 & 47.2 & 62.2\\
\quad SelfElicit & 30.8 & 33.1 & 1.9 & 14.2 & 67.0 & 18.8 & 9.0 & 27.4\\
\addlinespace[3pt]
\multicolumn{9}{l}{\textit{Jeff-1.1-Qwen-4B}}\\
\quad \method{} (ours) & \textbf{83.1} & \textbf{81.4} & \textbf{87.4} & \textbf{90.8} & \textbf{95.1} & \textbf{94.7} & \textbf{85.8} & \textbf{88.8}\\
\quad Training-free sum (ours) & {35.8} & {40.4} & {27.2} & {56.5} & {34.5} & {75.1} & {33.0} & {45.3}\\
\noalign{\vskip 1pt}\multispan{9}{\color{black!60}\dotfill}\cr\noalign{\vskip 1pt}
\quad {Attention (T)} & {43.3} & {59.3} & {55.8} & {73.6} & {65.2} & {91.3} & {59.1} & {64.9}\\
\quad AT2 & 63.4 & 64.7 & 72.3 & 86.1 & 75.9 & 89.6 & 75.3 & 76.2\\
\quad Leave-one-out & 75.9 & 63.8 & 76.7 & 82.3 & 80.9 & 78.4 & 76.0 & 77.1\\
\quad ContextCite-128 & 76.3 & 68.6 & 79.1 & 85.8 & 82.6 & 91.1 & 84.0 & 81.9\\
\quad LIME-32 & 72.2 & 65.3 & 76.2 & 72.1 & 75.1 & 80.1 & 70.1 & 73.2\\
\quad Kernel SHAP-32 & 63.5 & 66.1 & 54.4 & 50.2 & 63.0 & 60.7 & 41.6 & 57.3\\
\quad SelfElicit & 64.5 & 57.1 & 30.1 & 41.3 & 84.9 & 63.8 & 62.0 & 60.3\\
\bottomrule\end{tabular}
}
\caption{{\textbf{Post-trained models Hit@1 per dataset.} The Hit@1 column of \cref{tab:all-methods-with-sum} split by dataset, with the same methods. \method{} scores average three training seeds. Bold marks the best method per dataset within each model.}}
\label{tab:all-methods-sets}
\end{table}
\begin{table}[H]\centering\footnotesize\setlength{\tabcolsep}{3.6pt}\renewcommand{\arraystretch}{0.92}
{
\begin{tabular}{lrrrrrrr@{\hspace{12pt}}r}\toprule
\emph{Base LLMs} & MultiRC & SciFact & ContractNLI & HotpotQA & FinQA & Evidence Inf. & ConditionalQA & All\\
Method & (598) & (354) & (206) & (598) & (551) & (527) & (445) & (3,279)\\\midrule
\multicolumn{9}{l}{\textit{Gemma-4-31B-it}}\\
\quad \method{} (ours) & \textbf{85.1} & \textbf{78.0} & \textbf{90.8} & \textbf{92.0} & \textbf{94.0} & \textbf{91.3} & \textbf{88.8} & \textbf{88.9}\\
\quad Training-free sum (ours) & {81.1} & {74.0} & {87.4} & {88.5} & {68.4} & {82.0} & {87.4} & {80.9}\\
\noalign{\vskip 1pt}\multispan{9}{\color{black!60}\dotfill}\cr\noalign{\vskip 1pt}
\quad {Attention (T)} & {21.9} & {31.6} & {13.1} & {8.2} & {11.8} & {23.0} & {8.5} & {16.6}\\
\quad AT2 & 24.4 & 48.3 & 30.6 & 41.0 & 46.8 & 65.7 & 29.9 & 41.5\\
\quad Leave-one-out & 69.1 & 59.3 & 69.9 & 69.9 & 75.7 & 38.5 & 58.2 & 62.9\\
\quad ContextCite-128 & 78.1 & 68.9 & 85.4 & 79.8 & 84.6 & 81.0 & 76.4 & 79.2\\
\quad LIME-32 & 72.9 & 61.3 & 76.7 & 64.4 & 76.0 & 54.6 & 60.0 & 66.2\\
\quad Kernel SHAP-32 & 64.2 & 66.4 & 65.5 & 43.1 & 61.9 & 46.1 & 42.5 & 54.4\\
\quad SelfElicit & 16.1 & 31.1 & 3.4 & 4.0 & 10.3 & 9.3 & 4.3 & 11.0\\
\addlinespace[3pt]
\multicolumn{9}{l}{\textit{Gemma-4-12B-it}}\\
\quad \method{} (ours) & \textbf{84.3} & \textbf{82.2} & \textbf{89.8} & 88.5 & \textbf{95.8} & \textbf{92.6} & \textbf{88.5} & \textbf{89.0}\\
\quad Training-free sum (ours) & {84.1} & {76.8} & {85.9} & {\textbf{88.6}} & {72.1} & {88.2} & {85.6} & {83.1}\\
\noalign{\vskip 1pt}\multispan{9}{\color{black!60}\dotfill}\cr\noalign{\vskip 1pt}
\quad {Attention (T)} & {21.9} & {32.8} & {15.5} & {16.6} & {12.0} & {24.7} & {16.4} & {19.7}\\
\quad AT2 & 34.8 & 53.4 & 61.2 & 47.3 & 69.3 & 80.5 & 58.2 & 57.1\\
\quad Leave-one-out & 63.7 & 61.6 & 77.2 & 68.7 & 57.9 & 57.3 & 68.8 & 63.9\\
\quad ContextCite-128 & 64.5 & 70.6 & 85.0 & 76.9 & 63.0 & 87.3 & 82.9 & 74.6\\
\quad LIME-32 & 61.7 & 65.3 & 80.6 & 60.7 & 54.8 & 68.9 & 69.0 & 64.1\\
\quad Kernel SHAP-32 & 52.3 & 67.2 & 61.2 & 36.3 & 37.9 & 48.8 & 36.2 & 46.4\\
\quad SelfElicit & 17.2 & 23.7 & 4.4 & 5.9 & 11.8 & 6.3 & 9.9 & 11.4\\
\addlinespace[3pt]
\multicolumn{9}{l}{\textit{Qwen3.6-27B}}\\
\quad \method{} (ours) & 76.4 & \textbf{76.0} & \textbf{86.9} & 74.6 & \textbf{85.3} & 76.5 & 74.4 & 77.9\\
\quad Training-free sum (ours) & {26.3} & {14.1} & {4.4} & {13.4} & {7.4} & {11.4} & {10.6} & {13.5}\\
\noalign{\vskip 1pt}\multispan{9}{\color{black!60}\dotfill}\cr\noalign{\vskip 1pt}
\quad {Attention (T)} & {25.8} & {16.1} & {2.9} & {18.2} & {16.7} & {12.7} & {10.1} & {16.2}\\
\quad AT2 & 26.9 & 45.8 & 24.3 & 46.5 & 73.0 & 62.6 & 23.4 & 45.3\\
\quad Leave-one-out & 76.3 & 61.9 & 81.1 & 79.1 & 70.4 & 71.7 & 71.7 & 73.2\\
\quad ContextCite-128 & \textbf{76.9} & 69.8 & 84.5 & \textbf{85.3} & 77.5 & \textbf{91.1} & \textbf{82.2} & \textbf{81.2}\\
\quad LIME-32 & 74.2 & 66.9 & 82.5 & 69.6 & 67.7 & 75.0 & 70.1 & 71.6\\
\quad Kernel SHAP-32 & 65.7 & 70.3 & 66.0 & 39.8 & 50.1 & 53.5 & 44.3 & 54.0\\
\quad SelfElicit & 22.7 & 21.2 & 3.9 & 5.2 & 40.3 & 3.8 & 4.0 & 15.6\\
\addlinespace[3pt]
\multicolumn{9}{l}{\textit{Qwen3.5-4B}}\\
\quad \method{} (ours) & 63.4 & \textbf{79.7} & 66.5 & \textbf{75.3} & \textbf{82.8} & 76.3 & 70.3 & \textbf{73.8}\\
\quad Training-free sum (ours) & {25.4} & {16.4} & {2.9} & {20.2} & {8.2} & {11.8} & {8.3} & {14.7}\\
\noalign{\vskip 1pt}\multispan{9}{\color{black!60}\dotfill}\cr\noalign{\vskip 1pt}
\quad {Attention (T)} & {25.4} & {20.3} & {1.9} & {24.2} & {10.7} & {21.3} & {8.8} & {17.8}\\
\quad AT2 & 31.4 & 48.0 & 17.0 & 46.2 & 53.9 & 53.9 & 25.4 & 41.6\\
\quad Leave-one-out & \textbf{64.2} & 64.4 & 74.8 & 68.7 & 61.0 & 67.4 & 69.0 & 66.3\\
\quad ContextCite-128 & 62.4 & 74.9 & \textbf{77.2} & 70.9 & 67.7 & \textbf{82.4} & \textbf{72.6} & 71.7\\
\quad LIME-32 & 61.0 & 69.5 & 74.3 & 54.5 & 60.4 & 70.6 & 59.8 & 62.9\\
\quad Kernel SHAP-32 & 51.0 & 66.4 & 43.7 & 30.3 & 32.3 & 44.2 & 31.0 & 41.5\\
\quad SelfElicit & 26.6 & 22.6 & 2.9 & 5.0 & 23.2 & 5.5 & 9.9 & 14.5\\
\addlinespace[3pt]
\multicolumn{9}{l}{\textit{Model-independent}}\\
\quad BM25 & 73.2 & 41.0 & 30.1 & 20.6 & 42.8 & 18.0 & 41.6 & 39.2\\
\quad Random & 20.4 & 21.8 & 2.4 & 5.5 & 6.4 & 5.1 & 5.8 & 9.9\\
\bottomrule\end{tabular}
}
\caption{{\textbf{Base LLMs Hit@1 per dataset.} As \cref{tab:all-methods-sets}, for the four base LLMs BM25 and Random are methods that do not depend on a System One model.}}
\label{tab:all-methods-sets-pp}
\end{table}

\FloatBarrier

\FloatBarrier
\subsection{Latency}\label{app:latency-mean}

\begin{figure}[!htbp]
\centering
\includegraphics[width=\linewidth]{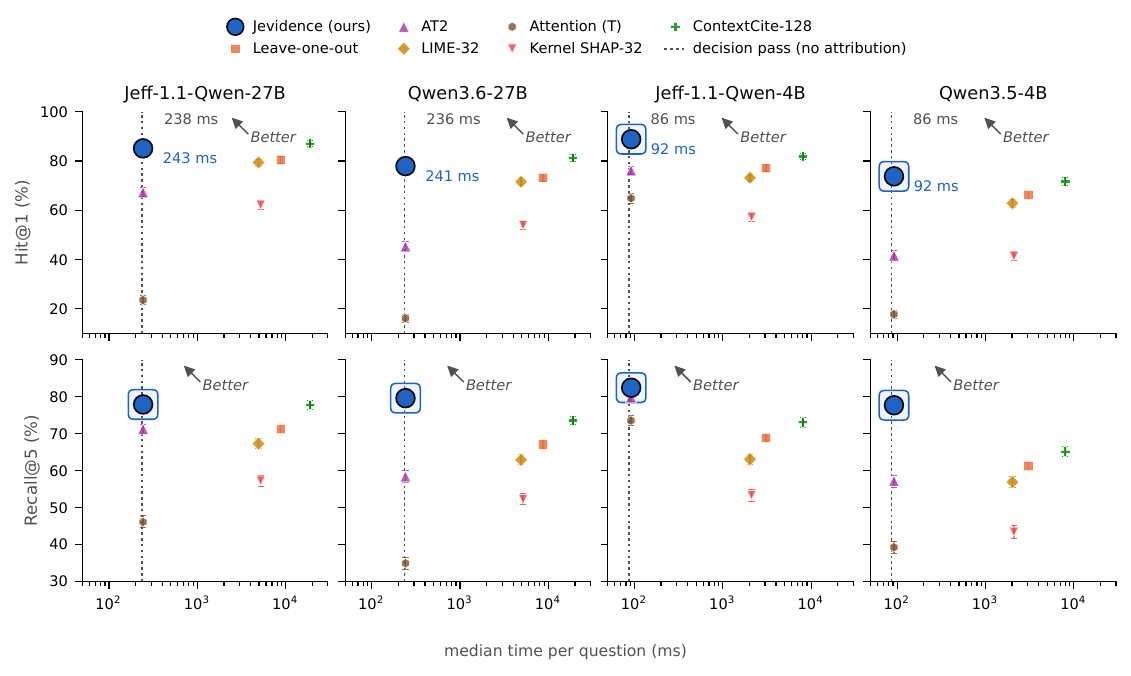}
\caption{{\textbf{Performance vs. inference latency.} Hit@1 (top) and Recall@5 (bottom) over all 3,279 marked questions against the median time per question. The dotted line is the decision pass without attribution. The large blue circle is the same pass with the additional overhead of attribution with \method{}. If there is a blue square this indicates the method dominates all methods on both performance and latency. The labels in the top row are the median times of the decision pass (grey) and \method (blue). }}
\label{fig:latency-qwen}
\end{figure}

\begin{figure}[!htbp]
\centering
\includegraphics[width=\linewidth]{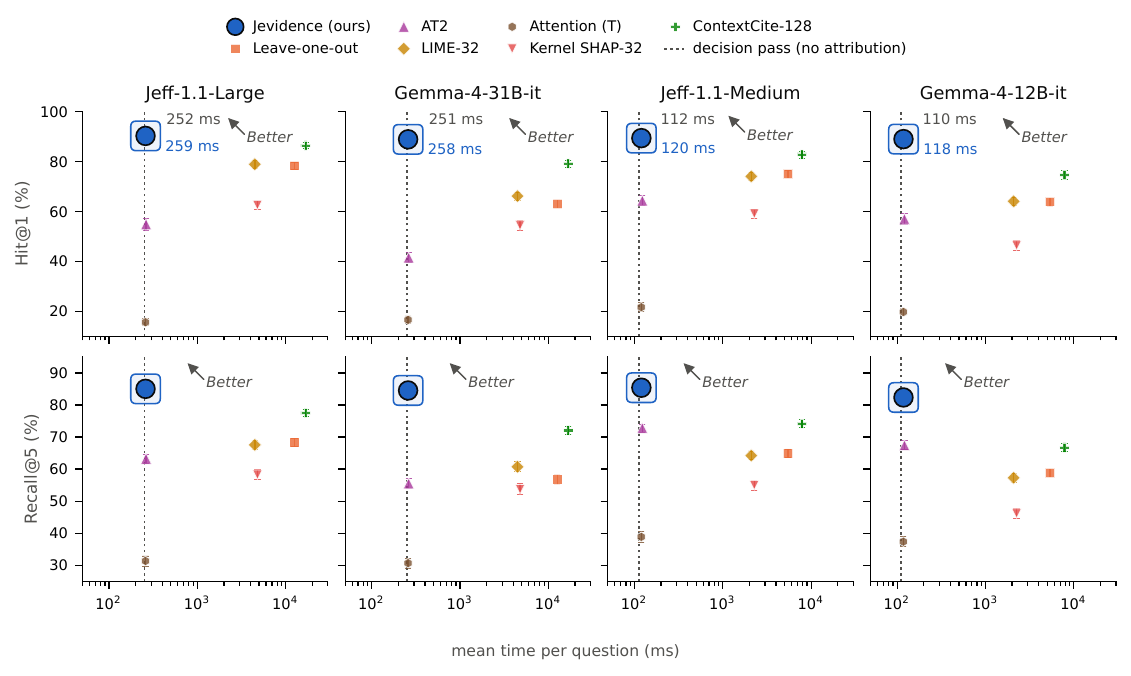}
The same plots as in \caption{{Figure~\ref{fig:latency} but with the mean instead of the median over the 3,279 marked questions. Leave-one-out moves from 34--36$\times$ to 45--49$\times$ \method{}'s time, because its number of passes grows with the number of units in the document. ContextCite-128, LIME-32 and Kernel SHAP-32 move slightly closer (their ratio to \method{}'s time is multiplied by 0.93--0.96). \method{}'s overhead over the decision pass grows from 2.2--5.2\% to 2.7--7.4\%. The labels in the top row are the mean times of the decision pass (grey) and \method{} (blue). As in Figure~\ref{fig:latency}, the dotted line is the decision pass without attribution, and a blue square indicates that the method dominates all methods on both performance and latency. Error bars are 95\% bootstrap intervals.}}
\label{fig:latency-mean}
\end{figure}

\begin{figure}[!htbp]
\centering
\includegraphics[width=\linewidth]{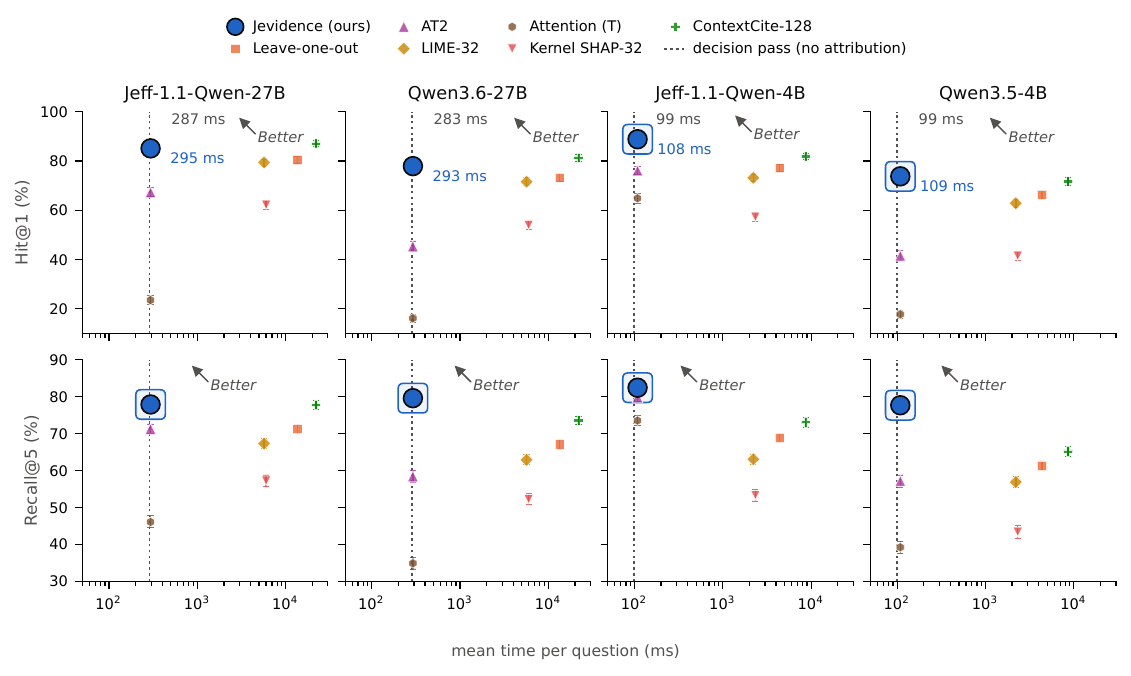}
\caption{The same plots as in {Figure~\ref{fig:latency-qwen} but with the mean instead of the median over the 3,279 marked questions and computed as in Figure~\ref{fig:latency-mean}. Leave-one-out moves from 33--36$\times$ to 40--46$\times$ \method{}'s time; ContextCite-128, LIME-32 and Kernel SHAP-32 move slightly closer (their ratio to \method{}'s time is multiplied by 0.91--0.96). \method{}'s overhead over the decision pass grows from 2.3--7.1\% to 3.0--9.4\%. The slowest 1\% of the questions account for at most 6.4\% of any method's total time. AT2's time is \method{}'s mean plus the difference of the two means on the 164-question timing subset ($-0.3$ to $-0.1$~ms), so its position is indicative; Attention (T)'s likewise, from the attention sum ($-0.2$ to $-0.1$~ms). The labels in the top row are the mean times of the decision pass (grey) and \method{} (blue). Error bars are 95\% bootstrap intervals.}}
\label{fig:latency-qwen-mean}
\end{figure}

Figure~\ref{fig:latency-qwen} repeats Figure~\ref{fig:latency} for the Qwen models. ContextCite-128's higher Hit@1 on the two Qwen-27B models comes at 78--88 times the latency of \method{}. Figures~\ref{fig:latency-mean} and~\ref{fig:latency-qwen-mean} repeat the two latency figures with the mean instead of the median time per question.

\FloatBarrier

\subsection{{Retrieval quality by document length}}\label{app:length}
\begin{figure}[!htbp]
\centering
\includegraphics[width=0.84\linewidth]{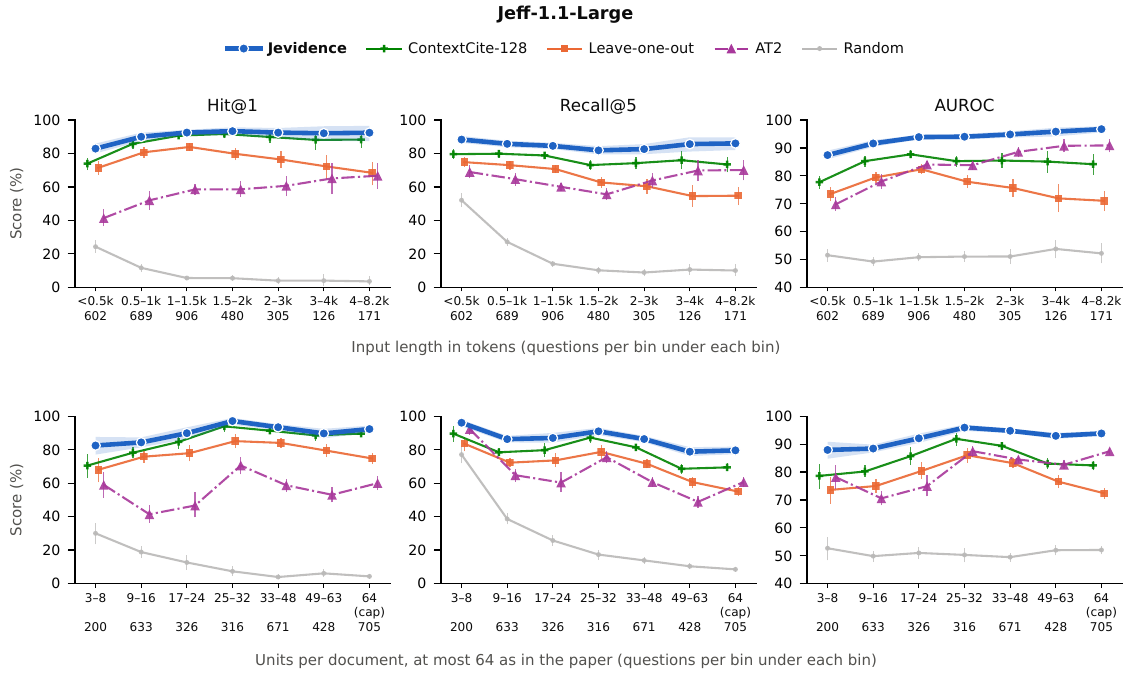}

\vspace{4pt}
\includegraphics[width=0.84\linewidth]{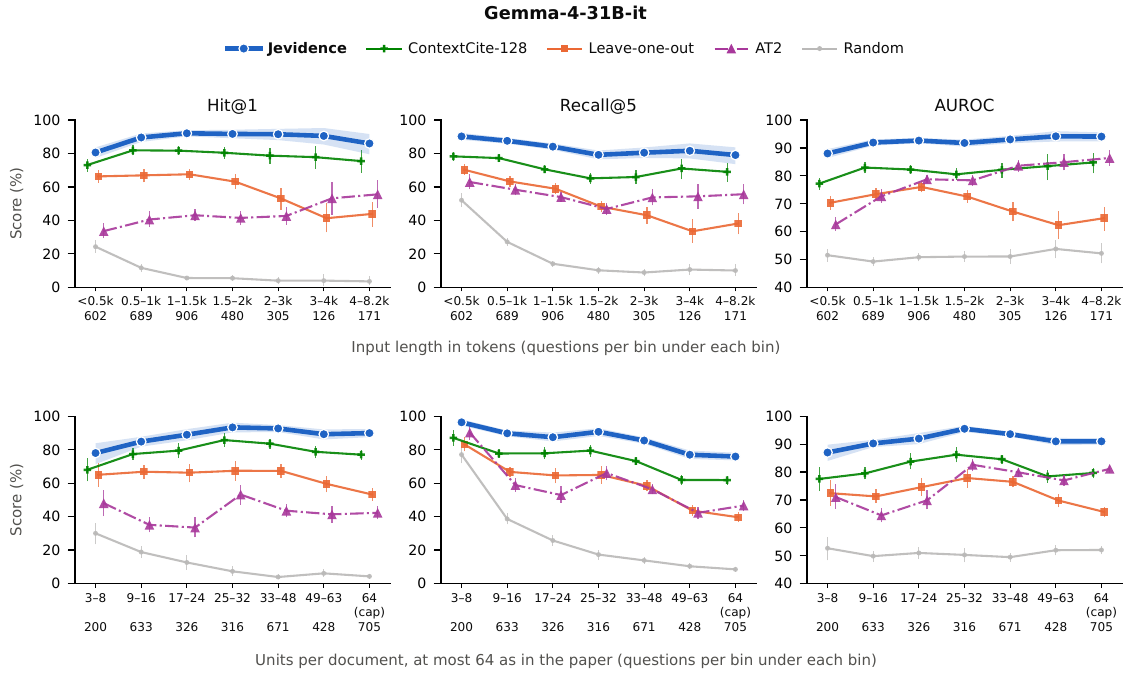}
\caption{{\textbf{Retrieval quality by input length and by number of units.} Top: Jeff-1.1-Large. Bottom: Gemma-4-31B-it. Columns: Hit@1, Recall@5 and AUROC (\%) on the 3,279 marked questions. For each model, the first row groups questions by the length of the model's full input in tokens, and the second by the number of units in their document. The number under each bin is its number of questions. The shaded band (\method{}) and the error bars are 95\% confidence intervals.}}
\label{fig:length}
\end{figure}

\method{} is trained on questions with much shorter context than the evaluation datasets (See Table~\ref{tab:input-size} and Figure~\ref{fig:length-dist}). Despite the mismatch in context size, we show that \method{}'s performance is robust across number of tokens and number of units on Gemma-4-31B-it and Jeff-1.1-Large, and generalizes well to the evaluation datasets. Figure~\ref{fig:length} shows that, under our segmentation rules, \method{} performs well even at the longest documents.

\begin{figure}[!htbp]
\centering
\includegraphics[width=0.9\linewidth]{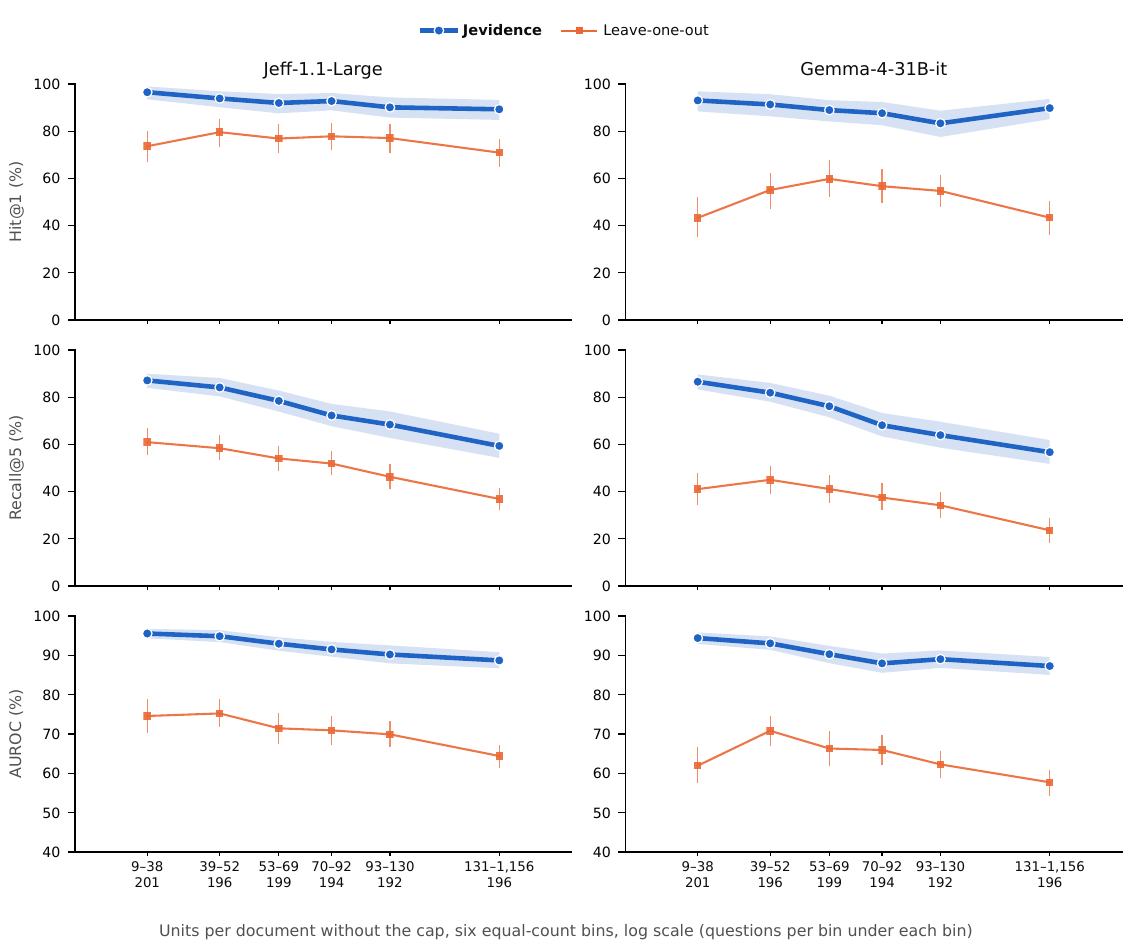}
\caption{{\textbf{Retrieval performance across units (without 64-unit cap).} Hit@1, Recall@5 and AUROC (\%) for Jeff-1.1-Large (left) and Gemma-4-31B-it (right) on the 1,178 marked questions of ContractNLI, Evidence Inference and ConditionalQA, rerun with every unit of the segmentation kept (up to 1,156 units per document). Questions are grouped into six bins of equal size by the number of units in their document (log scale); the number under each bin is its number of questions.}}
\label{fig:length-nomerge}
\end{figure}

However, our segmentation rules cap the number of units in long documents to be 64 units (Appendix~\ref{app:segmentation}) by iteratively merging short adjacent units. A natural question to ask is what happens if we allow for an arbitrary number of units? To answer this, in Figure~\ref{fig:length-nomerge} we compare \method{} and LOO across documents without the 64 unit cap. In Figure~\ref{fig:length-nomerge}, we restrict to ConditionalQA, Evidence Inference, and ContractNLI, since they contain 93\% of the documents with more than 64 units . Our results indicate that removing the 64-unit cap does not change the performance gap between \method{} and leave-one-out. Both methods' Hit@1 and AUROC exhibit mild decline; their Recall@5 declines more significantly, which is explained in part by the fact that with more units, more documents have more than five units total, which leaves an irreducible Recall@5 gap. We leave training \method{} on longer documents with more units for future work.

\FloatBarrier

\subsection{Comparing \method{} to leave-one-out}
\label{app:rd-split}

Although \method{} is trained exclusively on leave-one-out
(LOO) targets, it achieves higher agreement with
human-annotated evidence than LOO itself
(\cref{sec:performance}). This might seem counterintuitive at first, and one might think that \method{}'s downstream performance should be upper bounded by the quality of its training targets. We investigate this difference
by examining both methods' performance across different categories of evaluation samples, which yields a clean explanation of the observed gap.

\paragraph{Experimental setup.}
We aim to categorize samples by what effect its marked units have on the model prediction. To this end, for each of the 3,279 evaluation questions, in addition to the forward pass on the complete context, which provides $p(\hat{a})$, the probability of the answer picked by the model, and the per-unit forward passes to compute every unit's LOO score,  we run two additional forward passes: deleting \textit{all} marked units, and
retaining \textit{only} the marked units.
We then partition questions into four mutually exclusive
categories, checked in order (meaning a decision can only be in a category if it is not contained in any of the preceding categories):

\begin{enumerate}[label=(\arabic*),start=1,nosep,leftmargin=2.4em]
\item \emph{Single mark moves $p(\hat{a})$:} There exists at least one
marked unit whose deleteion reduces $p(\hat a)$ by at least 0.05.
\item \emph{Only all marked units move $p(\hat{a})$:} No individual marked unit
meets this threshold, but removing all marked units
together does, with at least two marked units present.
\item \emph{Evidence in non-marked units:} Removing all
marked units changes $p(\hat a)$ by less than 0.05,
yet retaining only the marked units preserves
$\hat a$ with at least half its original probability,
and removing the entire document changes the
predicted answer.
\item \emph{Decision does not need the document:} All remaining questions. These questions are essentially those where the model does not require the document to answer the question correctly.
\end{enumerate}

\begin{figure}[H]
\centering
\includegraphics[width=\linewidth]{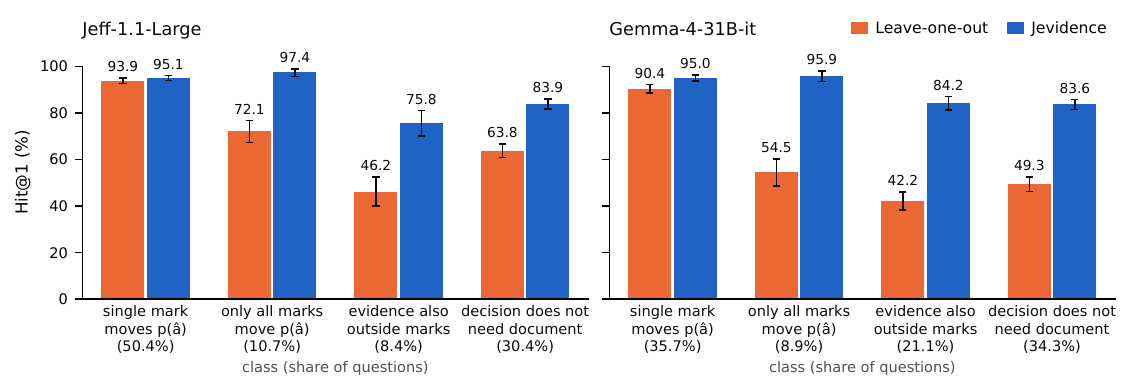}

\caption{\textbf{\method{}'s advantage over LOO is
largest when individual marked-unit deletions have
small effects.}
Hit@1 of LOO (orange) and \method{} (blue) on
Jeff-1.1-Large (left) and Gemma-4-31B-it (right), across four mutually exclusive question categories: (1) a single unit moves the original answer probability; (2) only all marked units together do; (3) evidence is contained in non-marked units as well; (4) the model does not need the context to answer the question. Percentages below category names are the shares amongst all evaluation questions, and error bars are 95\% confidence intervals.}
\label{fig:rd-split}
\end{figure}

\paragraph{Results.}
\Cref{fig:rd-split} reports Hit@1 within each category
on Jeff-1.1-Large and Gemma-4-31B-it.
\method{}'s advantage is by far the smallest on questions where a single mark moves the probability (category 1); on Jeff-1.1-Large it is ahead by just 1.2\%, and on
Gemma-4-31B-it by 4.6\%. These questions are exactly those where LOO can detect the evidence.

The differences are considerably larger in categories
2--4, where no individual marked unit moves the prediction. Across these categories, \method{}
exceeds LOO by 20--30 percentage points on
Jeff-1.1-Large and 34--42 points on Gemma-4-31B-it.

\paragraph{Interpretation.}
These findings show that LOO scores can find marked evidence exactly on the questions where a single marked unit moves the model's prediction, which is, by definition of LOO, precisely the type of evidence LOO can detect. However, on questions where that is not the case -- if evidence is redundant, or contained in non-marked units as well, or the model does not rely on the context to answer the question correctly -- LOO collapses, while \method{} still faithfully recovers marked evidence. This shows that \method{} goes beyond its training targets, and works even where LOO doesn't.

\subsection{{Wrong decisions: Recall@5 and AUROC}}\label{app:wrong-decisions}

\begin{figure}[ht]
\centering
\includegraphics[width=\linewidth]{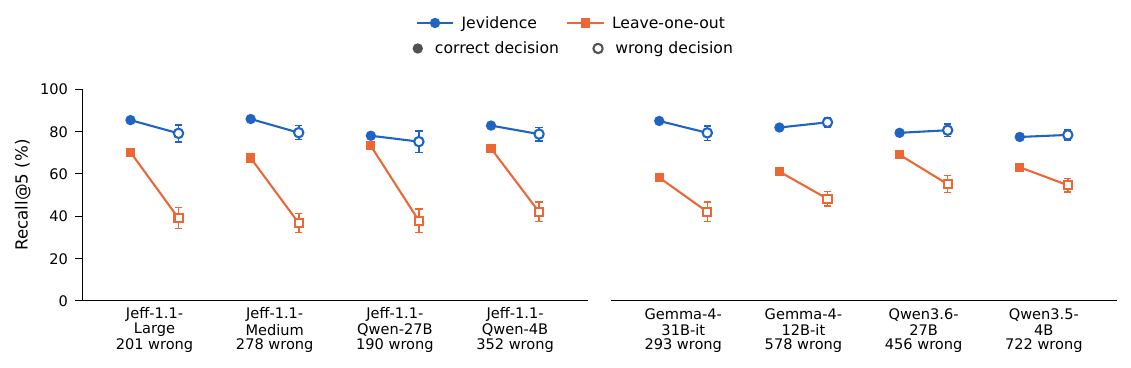}

\includegraphics[width=\linewidth]{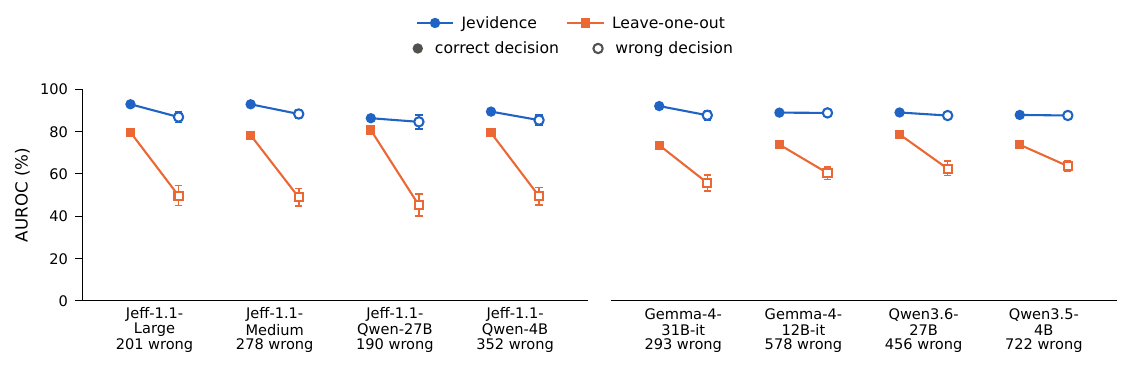}
\caption{{\textbf{Evidence retrieval when the model's decision is wrong: Recall@5 and AUROC.} Recall@5 (top) and AUROC (bottom) of \method{} (blue) and leave-one-out (orange) on the 3,279 marked questions, split by whether the model's decision matches the dataset's answer: correct (filled marker) or wrong (open marker). The number under each model is how many of the questions it answers wrongly. Error bars are 95\% intervals from 2,000 bootstrap resamples of document clusters.}}
\label{fig:wrong-decisions-metrics}
\end{figure}

{Figure~\ref{fig:wrong-decisions-metrics} repeats the comparison of Figure~\ref{fig:wrong-decisions} for Recall@5 and AUROC.} In table~\ref{tab:wrong-decisions-per-dataset} we show \method{}'s retrieval performance on correct/incorrect decisions on a per-dataset basis across 8 models.

\begin{table}[ht]\centering\footnotesize\setlength{\tabcolsep}{2.5pt}\renewcommand{\arraystretch}{0.92}
{
\begin{tabular}{@{}lrr@{\hspace{10pt}}rr@{\hspace{10pt}}rr@{\hspace{10pt}}rr@{\hspace{10pt}}rr@{\hspace{10pt}}rr@{\hspace{10pt}}rr@{\hspace{10pt}}rr@{\hspace{3pt}}}\toprule
 & \multicolumn{8}{c}{Jeff-1.1 post-trained models} & \multicolumn{8}{c}{Base LLMs}\\
\cmidrule(lr){2-9}\cmidrule(l){10-17}
 & \multicolumn{2}{c}{Large} & \multicolumn{2}{c}{Medium} & \multicolumn{2}{c}{Qwen-} & \multicolumn{2}{c}{Qwen-} & \multicolumn{2}{c}{Gemma-4-} & \multicolumn{2}{c}{Gemma-4-} & \multicolumn{2}{c}{Qwen3.6-} & \multicolumn{2}{c}{Qwen3.5-}\\
 & \multicolumn{2}{c}{} & \multicolumn{2}{c}{} & \multicolumn{2}{c}{27B} & \multicolumn{2}{c}{4B} & \multicolumn{2}{c}{31B-it} & \multicolumn{2}{c}{12B-it} & \multicolumn{2}{c}{27B} & \multicolumn{2}{c}{4B}\\
\cmidrule(lr){2-3}\cmidrule(lr){4-5}\cmidrule(lr){6-7}\cmidrule(lr){8-9}\cmidrule(lr){10-11}\cmidrule(lr){12-13}\cmidrule(lr){14-15}\cmidrule(l){16-17}
Dataset ($n$) & C & W & C & W & C & W & C & W & C & W & C & W & C & W & C & W\\
\midrule
\multicolumn{17}{@{}l}{\textbf{Wrong decisions}}\\
MultiRC (598) & \multicolumn{2}{c}{42} & \multicolumn{2}{c}{41} & \multicolumn{2}{c}{42} & \multicolumn{2}{c}{52} & \multicolumn{2}{c}{45} & \multicolumn{2}{c}{41} & \multicolumn{2}{c}{50} & \multicolumn{2}{c}{71}\\
SciFact (354) & \multicolumn{2}{c}{44} & \multicolumn{2}{c}{54} & \multicolumn{2}{c}{44} & \multicolumn{2}{c}{62} & \multicolumn{2}{c}{48} & \multicolumn{2}{c}{56} & \multicolumn{2}{c}{49} & \multicolumn{2}{c}{38}\\
ContractNLI (206) & \multicolumn{2}{c}{27} & \multicolumn{2}{c}{31} & \multicolumn{2}{c}{31} & \multicolumn{2}{c}{38} & \multicolumn{2}{c}{19} & \multicolumn{2}{c}{30} & \multicolumn{2}{c}{26} & \multicolumn{2}{c}{40}\\
HotpotQA (598) & \multicolumn{2}{c}{17} & \multicolumn{2}{c}{28} & \multicolumn{2}{c}{13} & \multicolumn{2}{c}{33} & \multicolumn{2}{c}{29} & \multicolumn{2}{c}{42} & \multicolumn{2}{c}{28} & \multicolumn{2}{c}{77}\\
FinQA (551) & \multicolumn{2}{c}{14} & \multicolumn{2}{c}{36} & \multicolumn{2}{c}{8} & \multicolumn{2}{c}{84} & \multicolumn{2}{c}{81} & \multicolumn{2}{c}{312} & \multicolumn{2}{c}{234} & \multicolumn{2}{c}{363}\\
Evidence Inf. (527) & \multicolumn{2}{c}{38} & \multicolumn{2}{c}{47} & \multicolumn{2}{c}{29} & \multicolumn{2}{c}{44} & \multicolumn{2}{c}{43} & \multicolumn{2}{c}{53} & \multicolumn{2}{c}{40} & \multicolumn{2}{c}{69}\\
ConditionalQA (445) & \multicolumn{2}{c}{19} & \multicolumn{2}{c}{41} & \multicolumn{2}{c}{23} & \multicolumn{2}{c}{39} & \multicolumn{2}{c}{28} & \multicolumn{2}{c}{44} & \multicolumn{2}{c}{29} & \multicolumn{2}{c}{64}\\
\addlinespace[1pt]
All (3,279) & \multicolumn{2}{c}{201} & \multicolumn{2}{c}{278} & \multicolumn{2}{c}{190} & \multicolumn{2}{c}{352} & \multicolumn{2}{c}{293} & \multicolumn{2}{c}{578} & \multicolumn{2}{c}{456} & \multicolumn{2}{c}{722}\\
\midrule
\multicolumn{17}{@{}l}{\textbf{Hit@1}}\\
MultiRC (598) & 85.3 & 73.8 & 84.9 & 75.6 & 83.1 & 73.8 & 84.1 & 73.1 & 85.7 & 77.8 & 85.1 & 73.2 & 76.8 & 72.0 & 65.5 & 47.9\\
SciFact (354) & 83.5 & 65.9 & 86.0 & 61.1 & 87.1 & 63.6 & 86.6 & 56.5 & 80.1 & 64.6 & 86.2 & 60.7 & 79.0 & 57.1 & 83.2 & 50.0\\
ContractNLI (206) & 95.0 & 70.4 & 93.1 & 58.1 & 91.4 & 61.3 & 91.7 & 68.4 & 94.7 & 52.6 & 94.3 & 63.3 & 91.7 & 53.8 & 70.5 & 50.0\\
HotpotQA (598) & 89.5 & 88.2 & 93.5 & 96.4 & 88.7 & 69.2 & 90.6 & 93.9 & 92.1 & 89.7 & 88.3 & 90.5 & 74.9 & 67.9 & 75.2 & 75.3\\
FinQA (551) & 97.2 & 85.7 & 92.4 & 83.3 & 69.2 & 75.0 & 95.1 & 95.2 & 94.9 & 88.9 & 96.2 & 95.5 & 85.8 & 84.6 & 83.5 & 82.4\\
Evidence Inf. (527) & 95.9 & 81.6 & 95.2 & 72.3 & 96.6 & 65.5 & 95.7 & 84.1 & 92.6 & 76.7 & 94.7 & 73.6 & 78.0 & 57.5 & 78.2 & 63.8\\
ConditionalQA (445) & 92.5 & 84.2 & 90.8 & 78.0 & 92.9 & 82.6 & 87.7 & 66.7 & 89.7 & 75.0 & 89.8 & 77.3 & 75.0 & 65.5 & 71.4 & 64.1\\
\addlinespace[1pt]
All (3,279) & 91.2 & 76.1 & 90.9 & 73.7 & 86.1 & 68.9 & 90.2 & 77.6 & 90.0 & 77.8 & 89.9 & 85.1 & 78.6 & 73.9 & 74.5 & 71.3\\
\midrule
\multicolumn{17}{@{}l}{\textbf{Recall@5}}\\
MultiRC (598) & 85.4 & 78.2 & 86.6 & 81.5 & 81.5 & 76.6 & 83.4 & 78.5 & 87.0 & 76.7 & 80.6 & 76.4 & 79.7 & 81.3 & 77.3 & 71.5\\
SciFact (354) & 91.4 & 82.6 & 93.0 & 87.3 & 94.5 & 87.1 & 91.0 & 85.2 & 95.9 & 86.7 & 90.0 & 84.4 & 91.8 & 80.8 & 91.5 & 73.9\\
ContractNLI (206) & 84.6 & 77.8 & 83.4 & 72.6 & 84.5 & 73.1 & 80.6 & 75.0 & 83.0 & 65.8 & 84.8 & 86.1 & 85.4 & 66.7 & 74.7 & 65.8\\
HotpotQA (598) & 81.6 & 81.9 & 87.1 & 83.9 & 75.0 & 66.7 & 77.6 & 77.3 & 81.8 & 76.7 & 77.5 & 72.8 & 70.7 & 70.5 & 72.6 & 72.1\\
FinQA (551) & 94.7 & 85.7 & 90.8 & 86.8 & 63.1 & 56.8 & 91.0 & 88.1 & 94.2 & 89.4 & 93.6 & 92.3 & 85.2 & 88.5 & 87.0 & 87.1\\
Evidence Inf. (527) & 89.8 & 82.9 & 88.3 & 78.8 & 89.6 & 74.2 & 90.6 & 80.6 & 87.0 & 76.2 & 86.5 & 73.1 & 84.5 & 65.9 & 83.5 & 76.9\\
ConditionalQA (445) & 69.8 & 60.6 & 70.2 & 63.6 & 68.3 & 65.5 & 65.7 & 51.8 & 67.1 & 59.1 & 70.1 & 58.8 & 68.8 & 58.3 & 61.9 & 56.6\\
\addlinespace[1pt]
All (3,279) & 85.4 & 79.2 & 85.9 & 79.5 & 78.0 & 75.2 & 82.8 & 78.8 & 85.0 & 79.4 & 81.9 & 84.4 & 79.4 & 80.6 & 77.4 & 78.4\\
\midrule
\multicolumn{17}{@{}l}{\textbf{AUROC}}\\
MultiRC (598) & 88.6 & 82.2 & 88.5 & 86.1 & 85.3 & 83.3 & 86.0 & 81.2 & 89.1 & 80.1 & 83.9 & 81.6 & 83.5 & 85.0 & 80.9 & 74.0\\
SciFact (354) & 88.5 & 79.8 & 89.4 & 81.2 & 91.5 & 81.5 & 86.9 & 79.0 & 90.5 & 80.2 & 86.9 & 75.3 & 86.6 & 76.5 & 88.6 & 71.5\\
ContractNLI (206) & 94.6 & 91.2 & 94.1 & 89.4 & 94.3 & 90.6 & 88.6 & 89.1 & 93.3 & 88.7 & 94.8 & 91.0 & 94.9 & 91.0 & 92.1 & 86.0\\
HotpotQA (598) & 93.1 & 90.7 & 94.5 & 90.6 & 89.1 & 92.3 & 87.6 & 89.5 & 92.2 & 87.4 & 87.6 & 84.9 & 89.3 & 90.4 & 87.4 & 88.7\\
FinQA (551) & 97.4 & 89.3 & 95.8 & 95.4 & 67.8 & 71.0 & 93.8 & 91.2 & 97.2 & 94.9 & 96.0 & 94.4 & 89.0 & 91.1 & 92.1 & 91.8\\
Evidence Inf. (527) & 97.5 & 94.4 & 96.7 & 90.7 & 97.0 & 83.4 & 96.8 & 91.5 & 96.1 & 93.2 & 94.1 & 82.2 & 94.4 & 84.2 & 94.8 & 93.2\\
ConditionalQA (445) & 90.0 & 87.0 & 90.8 & 88.5 & 87.8 & 86.5 & 85.0 & 74.7 & 85.8 & 83.1 & 86.4 & 83.5 & 88.9 & 80.9 & 85.2 & 81.8\\
\addlinespace[1pt]
All (3,279) & 92.9 & 86.9 & 92.9 & 88.3 & 86.3 & 84.6 & 89.4 & 85.4 & 92.1 & 87.7 & 89.0 & 88.8 & 89.0 & 87.6 & 87.9 & 87.6\\
\bottomrule\end{tabular}
}
\caption{{\textbf{\method{} by dataset when the model's decision is correct (C) or wrong (W).} Hit@1, Recall@5 and AUROC (\%) of \method{} on the $n$ marked questions of each dataset, split by whether the model's decision matches the dataset's answer; the top panel counts the wrong decisions, and All pools the seven datasets.}}
\label{tab:wrong-decisions-per-dataset}
\end{table}

\subsection{Wrong answer proposals across methods}\label{app:finqa-cf-methods}

\begin{figure}[ht]
\centering
\includegraphics[width=\linewidth]{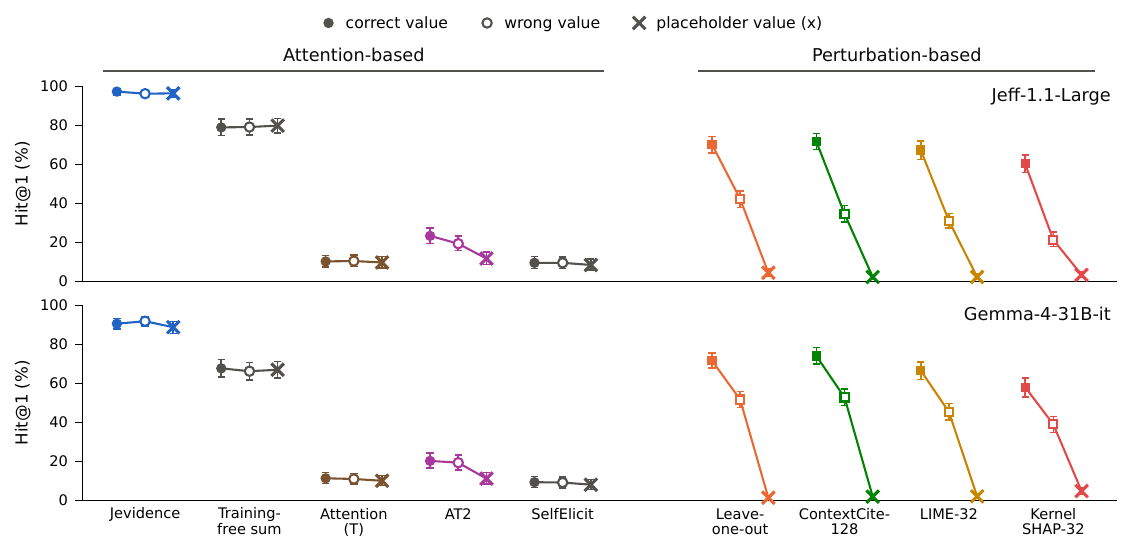}
\caption{\textbf{Wrong proposals hurt perturbation methods far more than attention methods.} We repeat the FinQA-CF experiment of \cref{sec:finqa-cf} across attention-based and perturbation-based models. Hit@1 on FinQA-CF for the five attention-based methods (circles, left) and the four perturbation-based methods (squares, right), each in its colour from Figure~\ref{fig:latency}; the training-free sum and SelfElicit, which that figure does not show, are in grey on Jeff-1.1-Large (top) and Gemma-4-31B-it (bottom). For each method we plot, from left to right, the three proposed values of \cref{sec:finqa-cf}: the correct value (filled marker), the wrong value (open marker), and the placeholder \texttt{x} ($\times$). For the placeholder, AT2 uses the weights it was fitted with, applied to the attention on the placeholder questions. The training-free sum is the version with $T$ (\cref{sec:training_free}).}
\label{fig:finqa-cf-methods}
\end{figure}

\Cref{fig:finqa-cf-methods} repeats the FinQA-CF comparison of \cref{sec:finqa-cf} with additional methods on Jeff-1.1-Large and Gemma-4-31B-it. We delineate the methods into two groups: (i) perturabtion methods; (ii) attention-based methods. The figure illustrates the relative robustness of the latter compared to the former.

\FloatBarrier

\subsection{\method{} versus Locate}\label{app:locate}

\paragraph{Locate's instructions.}
Using Perplexity's Locate requires an independent forward pass to retrieve evidence. In particular, Locate requires specific instructions or descriptions on what to find in the input context. It was trained on extractive question answering without multiple-choice options. To adapt it to our multiple-choice setting, we tried different prompt instructions (Table~\ref{tab:locate-prompts}), as well as either including or removing the set of answer options in the input. For each model, we choose the best prompt variant based off of Locate's performance across the pooled evaluation questions.
Specifically, we report results under the \texttt{Our instruction + options} prompt for Decider v1.3 (27B) and the \texttt{Our instruction} prompt for Decider Flash v1.3 (4B){; Table~\ref{tab:locate-prompt-results} gives the results for every instruction}.

\begin{table}[H]\centering\footnotesize\setlength{\tabcolsep}{4pt}\renewcommand{\arraystretch}{1.15}
{%
\begin{tabular}{@{}l>{\raggedright\arraybackslash}p{9.6cm}c@{}}\toprule
Name & Instruction & Options line\\\midrule
Question only & \emph{question} & no\\
Our instruction & Evidence in the context for deciding this question.\newline\emph{question} & no\\
Our instruction + options & Evidence in the context for deciding this question.\newline\emph{question}\newline Options: \emph{options} & yes\\
Alternative & Extract all evidence/counterfacts/etc.\ in the context that answers/disproves this question: \emph{question} & no\\
Alternative + options & Extract all evidence/counterfacts/etc.\ in the context that answers/disproves this question: \emph{question}\newline Options: \emph{options} & yes\\
Alternative, short & Extract all evidence in the context that answers this question: \emph{question} & no\\
Alternative, short + options & Extract all evidence in the context that answers this question: \emph{question}\newline Options: \emph{options} & yes\\
\bottomrule\end{tabular}}
\caption{{\textbf{The seven Locate instructions.} We provide the exact instructions that we give Locate, with line breaks as shown. \emph{question} stands for the question, and \emph{options} for the answer options (each option's label and text, separated by semicolons). Locate receives the full document as the \emph{context} that the instruction refers to. \textit{Alternative} focuses on extracting the specific nature of evidence from our tasks.}}
\label{tab:locate-prompts}
\end{table}

\begin{table}[H]\centering\footnotesize\setlength{\tabcolsep}{5pt}\renewcommand{\arraystretch}{0.95}
{%
\begin{tabular}{@{}lccc@{\hspace{16pt}}ccc@{}}\toprule
 & \multicolumn{3}{c}{Decider v1.3 (27B)} & \multicolumn{3}{c}{Decider Flash v1.3 (4B)}\\
\cmidrule(lr){2-4}\cmidrule(l){5-7}
Instruction given to Locate & Hit@1 & Recall@5 & AUROC & Hit@1 & Recall@5 & AUROC\\\midrule
Question only & 81.7 & 83.4 & 92.5 & 79.7 & 83.7 & 93.1\\
\textbf{Our instruction} & 90.4 & 88.1 & \textbf{94.3} & \textbf{85.4} & \textbf{87.3} & \textbf{94.2}\\
\textbf{Our instruction + options} & \textbf{90.9} & \textbf{88.3} & 94.1 & 83.5 & 86.6 & 93.5\\
Suggested & 88.2 & 87.2 & 94.0 & 84.1 & 86.2 & 93.7\\
Suggested + options & 89.5 & 87.8 & 94.0 & 80.6 & 84.8 & 92.7\\
Suggested, short & 86.6 & 86.9 & 93.9 & 84.4 & 86.8 & \textbf{94.2}\\
Suggested, short + options & 87.6 & 87.6 & 93.9 & 83.0 & 86.2 & 93.4\\
\midrule
\method{} & 91.3 & 86.0 & 92.2 & 89.2 & 86.6 & 93.0\\
\bottomrule\end{tabular}}
\caption{{\textbf{Locate with each instruction} of Table~\ref{tab:locate-prompts}: Hit@1, Recall@5 and AUROC (\%) over all 3,279 questions. Bold numbers mark the best instruction in each column (both when they are equal after rounding). Bold names mark the two instructions we select, the one with the highest Hit@1 for each model: \emph{Our instruction + options} for Decider v1.3 and \emph{Our instruction} for Decider Flash v1.3. Tables~\ref{tab:locate} and~\ref{tab:locate-sets} and Figure~\ref{fig:locate-latency} report Locate with the selected instruction. The last row gives \method{} on the same questions.}}
\label{tab:locate-prompt-results}
\end{table}

First, Table~\ref{tab:locate} gives the pooled comparison of Section~\ref{sec:locate} while Table~\ref{tab:locate-sets} compares \method{} and Perplexity's Locate on a per-dataset basis across the two Decider models (Section~\ref{sec:locate}). In summary, on Decider v1.3, the overall Hit@1 of the two methods does
not differ, while Locate has the higher Recall@5 and
AUROC. For Decider Flash v1.3, \method{} has the higher Hit@1, with clear leads on three datasets (HotpotQA, FinQA and Evidence Inference). We note that Locate has been trained on the train split of HotPotQA, unlike \method{}.

\begin{table}[H]\centering\footnotesize\setlength{\tabcolsep}{4pt}\renewcommand{\arraystretch}{0.92}
{
\begin{tabular}{llrrrr}\toprule
Model & Method & Hit@1 & Recall@5 & AUROC & Latency (ms)\\\midrule
Decider v1.3 (27B) & \method{} & 91.3 & 86.0 & 92.2 & 236\\
 & Locate + Decision & 90.9 & 88.3 & 94.1 & 529\\
\midrule
Decider Flash v1.3 (4B) & \method{} & 89.2 & 86.6 & 93.0 & 92\\
 & Locate + Decision & 85.4 & 87.3 & 94.2 & 235\\
\bottomrule\end{tabular}
}

\caption{\textbf{\method{} and Locate on the two Decider models}, pooled across all 3,279 evaluation questions with  Hit@1, Recall@5 and AUROC. Latency: median time per question for both answer and its evidence. For Locate, we report the best result over the seven instructions we tried (Table~\ref{tab:locate-prompt-results}).}
\label{tab:locate}
\end{table}

\begin{table}[H]\centering\footnotesize\setlength{\tabcolsep}{4pt}\renewcommand{\arraystretch}{0.92}
{
\begin{tabular}{lrrrrrrr@{\hspace{10pt}}rrrrrrr}\toprule
 & \multicolumn{7}{c}{Decider v1.3 (27B)} & \multicolumn{7}{c}{Decider Flash v1.3 (4B)}\\
\cmidrule(lr){2-8}\cmidrule(lr){9-15}
 & Accuracy & \multicolumn{2}{c}{Hit@1} & \multicolumn{2}{c}{Recall@5} & \multicolumn{2}{c}{AUROC} & Accuracy & \multicolumn{2}{c}{Hit@1} & \multicolumn{2}{c}{Recall@5} & \multicolumn{2}{c}{AUROC}\\
\cmidrule(lr){3-4}\cmidrule(lr){5-6}\cmidrule(lr){7-8}\cmidrule(lr){10-11}\cmidrule(lr){12-13}\cmidrule(lr){14-15}
Dataset ($n$) & & J & L & J & L & J & L & & J & L & J & L & J & L\\\midrule
MultiRC (598) & 94.1 & 82.3 & \textbf{87.1} & 85.9 & \textbf{90.1} & 88.5 & \textbf{91.5} & 93.5 & \textbf{84.4} & 80.9 & 87.9 & \textbf{88.5} & 89.4 & \textbf{90.0}\\
SciFact (354) & 87.9 & \textbf{82.5} & 79.4 & 93.7 & \textbf{94.3} & \textbf{90.5} & 89.4 & 87.6 & 78.0 & \textbf{82.8} & 94.6 & \textbf{95.8} & 89.3 & \textbf{91.1}\\
ContractNLI (206) & 89.8 & \textbf{90.3} & \textbf{90.3} & \textbf{85.4} & 80.9 & \textbf{95.0} & 89.1 & 88.8 & 86.4 & \textbf{90.3} & 84.6 & \textbf{86.4} & \textbf{95.4} & \textbf{95.3}\\
HotpotQA (598) & 98.3 & \textbf{94.1} & 90.1 & 87.5 & \textbf{92.1} & 94.4 & \textbf{97.8} & 96.3 & \textbf{91.0} & 77.4 & 83.7 & \textbf{87.2} & 91.8 & \textbf{95.9}\\
FinQA (551) & 99.3 & \textbf{98.0} & 96.0 & 85.9 & \textbf{93.7} & 88.8 & \textbf{96.7} & 93.1 & \textbf{97.6} & 94.6 & \textbf{94.4} & 93.2 & \textbf{97.5} & \textbf{97.1}\\
Evidence Inf. (527) & 94.3 & 94.3 & \textbf{95.8} & \textbf{89.4} & 88.6 & \textbf{96.8} & \textbf{97.2} & 93.7 & \textbf{94.5} & 90.1 & \textbf{89.9} & 86.4 & \textbf{97.2} & 96.3\\
ConditionalQA (445) & 95.5 & \textbf{95.3} & 94.2 & \textbf{74.0} & 72.1 & \textbf{92.8} & 91.9 & 90.3 & \textbf{86.5} & 85.2 & 70.0 & \textbf{73.1} & 90.4 & \textbf{93.1}\\
\midrule
All (3,279) & 95.0 & \textbf{91.3} & \textbf{90.9} & 86.0 & \textbf{88.3} & 92.2 & \textbf{94.1} & 92.6 & \textbf{89.2} & 85.4 & 86.6 & \textbf{87.3} & 93.0 & \textbf{94.2}\\
\bottomrule\end{tabular}
}
\caption{\textbf{\method{} (J) and Locate (L) by dataset.} We provide the breakdown of performance across the evaluation dataset for \cref{tab:locate}. Accuracy is the percentage of questions that the Decider models answer correctly. Bold marks the larger value of each pair, or both when they differ by less than 0.5 points. {Locate uses the instruction selected for each model (Table~\ref{tab:locate-prompt-results}). We note that Locate has been trained on the training split of HotPotQA, unlike \method{}.}}
\label{tab:locate-sets}
\end{table}

\begin{figure}[H]
\centering
\includegraphics[width=\linewidth]{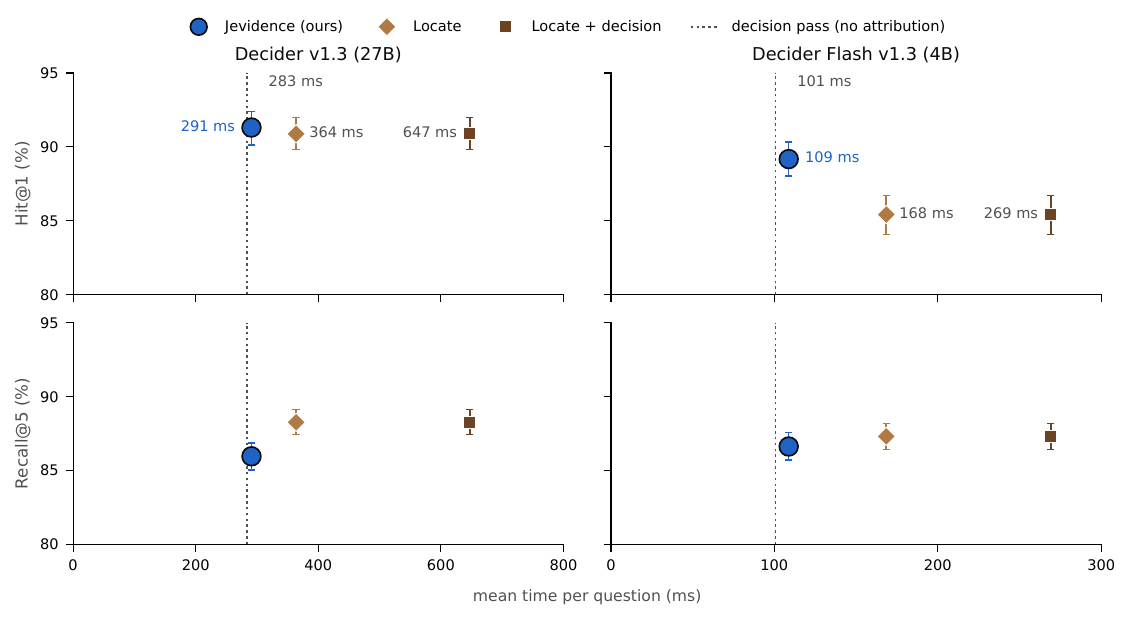}
\caption{{\cref{fig:locate-latency} with the mean instead of the median over the 3,279 marked questions: Hit@1 (top) and Recall@5 (bottom) against each question's time (median of five repetitions; for Locate + decision, the sum of the two passes) averaged over the same questions. The dotted  Error bars are 95\% confidence intervals. The dotted line is the inference latency of the decision pass alone.}}
\label{fig:locate-latency-mean}
\end{figure}

\subsection{Fitted coefficients of \method{}}\label{app:reader-weights}
\begin{figure}[ht]
\centering
\includegraphics[width=\linewidth]{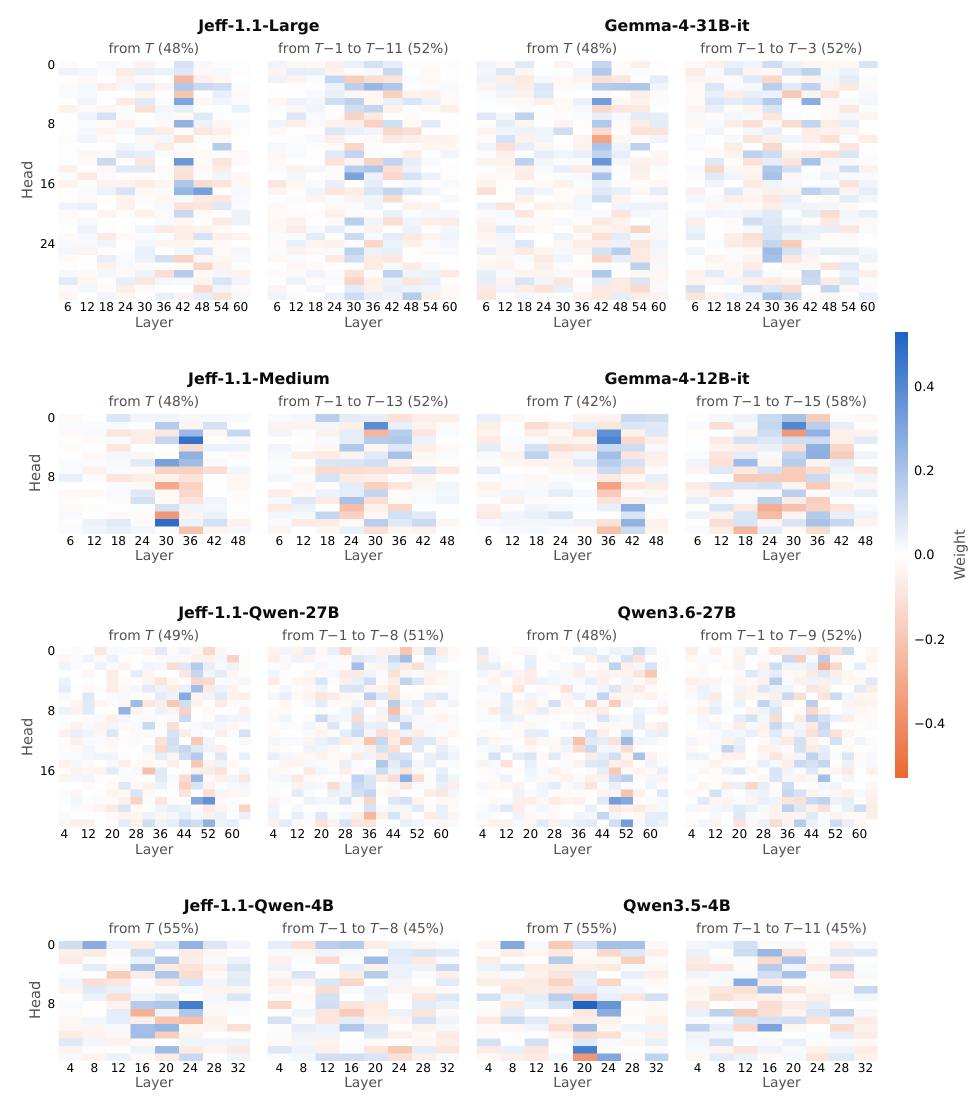}
\caption{\textbf{What \method{} learns.} The weights $\theta$ of \method{} for the four Jeff-1.1 post-trained models and their base LLMs. For each model, the left panel shows the weights on the attention from $T$ and the right panel those on the mean attention over $T-1$ to $T-K$, with the global-attention layers across and the heads down. Color is the weight on a head's attention divided by its root mean square over the training units, on one scale for all panels. The percentages are each position's share of the model's total absolute weight.}
\label{fig:reader-weights}
\end{figure}

Figure~\ref{fig:reader-weights} shows the weights $\theta$ that \method{} learns for each of the eight models. There is one weight per head and read position, one for the attention from $T$ and one for the mean attention over the $K$ positions before $T$ (Equation~\eqref{eq:attention-features}). The features are divided by their root mean square over the training units (\appref{app:reader}), so the weights are comparable across heads. The weights show three things: First, without enforcing this in any way at training, \method{} tends to use both read positions about equally (attention from $T$ receives 42--55\% of the total absolute weight, and preceding positions about 45--58\%). Second, 43--47\% of the absolute weight is on negative weights, so \method{} subtracts the attention of some heads instead of only adding up attention. This indicates that some head's attention points \textit{opposite} to the evidence. Third, the weight is spread over many heads, since the 10\% of heads with the largest weights hold only 27--36\% of it, and it is larger in the middle layers than in the first and last ones. This surprised us: we expected most of the weight to be concentrated in later layers. However, it seems that much of the important information for evidence attribution is built in earlier layers. We leave investigating this interesting phenomenon for future research.

\FloatBarrier

\clearpage
\section{Example outputs}\label{app:example-outputs}

We show \method{} on five questions from each of the seven evaluation datasets
(Figures~\ref{fig:ex-multirc-1}--\ref{fig:ex-condqa-5}), all with the  Jeff-1.1-Large post-trained model~\cite{opendecision2026}. Each figure shows the model's context input and its output which is its calibrated decision and retrieved evidence (which we restrict to top-3 units).

\newcommand{\jevexample}[3]{%
  \begin{center}
    \captionsetup{type=figure}%
    \includegraphics[width=\linewidth]{figures/examples/jev_example_#1_#2.pdf}
    \caption{#3, example #2 of 5.}\label{fig:ex-#1-#2}
  \end{center}
  \clearpage}

\jevexample{multirc}{1}{MultiRC}
\jevexample{multirc}{2}{MultiRC}
\jevexample{multirc}{3}{MultiRC}
\jevexample{multirc}{4}{MultiRC}
\jevexample{multirc}{5}{MultiRC}
\jevexample{scifact}{1}{SciFact}
\jevexample{scifact}{2}{SciFact}
\jevexample{scifact}{3}{SciFact}
\jevexample{scifact}{4}{SciFact}
\jevexample{scifact}{5}{SciFact}
\jevexample{contractnli}{1}{ContractNLI}
\jevexample{contractnli}{2}{ContractNLI}
\jevexample{contractnli}{3}{ContractNLI}
\jevexample{contractnli}{4}{ContractNLI}
\jevexample{contractnli}{5}{ContractNLI}
\jevexample{hotpotqa}{1}{HotpotQA}
\jevexample{hotpotqa}{2}{HotpotQA}
\jevexample{hotpotqa}{3}{HotpotQA}
\jevexample{hotpotqa}{4}{HotpotQA}
\jevexample{hotpotqa}{5}{HotpotQA}
\jevexample{finqa}{1}{FinQA}
\jevexample{finqa}{2}{FinQA}
\jevexample{finqa}{3}{FinQA}
\jevexample{finqa}{4}{FinQA}
\jevexample{finqa}{5}{FinQA}
\jevexample{evinf}{1}{Evidence Inference}
\jevexample{evinf}{2}{Evidence Inference}
\jevexample{evinf}{3}{Evidence Inference}
\jevexample{evinf}{4}{Evidence Inference}
\jevexample{evinf}{5}{Evidence Inference}
\jevexample{condqa}{1}{ConditionalQA}
\jevexample{condqa}{2}{ConditionalQA}
\jevexample{condqa}{3}{ConditionalQA}
\jevexample{condqa}{4}{ConditionalQA}
\jevexample{condqa}{5}{ConditionalQA}

\end{document}